\documentclass[11pt]{article}
\pdfoutput=1
\usepackage[
  margin=1.5cm,
  includefoot,
  footskip=30pt,
]{geometry}
\usepackage{graphicx} 
\usepackage{lipsum}
\usepackage[english, latin]{babel}
\usepackage{blindtext}
\usepackage{natbib}
\usepackage{booktabs}
\usepackage{multirow}
\usepackage[table,xcdraw]{xcolor}
\usepackage{array}
\usepackage{longtable}
\usepackage{hyperref}
\usepackage{caption}
\usepackage{dirtree}
\usepackage{amsmath}
\usepackage{amssymb}
\usepackage[normalem]{ulem}
\usepackage{subcaption}
\usepackage[export]{adjustbox}
\usepackage{multirow}
\usepackage{multicol}
\usepackage{scalerel} 
\usepackage{wrapfig}

\hypersetup{
  colorlinks,
  linkcolor={green!30!black},
  citecolor={blue!50!black},
  urlcolor={blue!50!black}
}

\title{Grounded Intuition of GPT-Vision's Abilities with Scientific Images}
\author{Alyssa Hwang \and Andrew Head \and Chris Callison-Burch}
\date{Department of Computer and Information Science \\ University of Pennsylvania}

\begin{document}

\definecolor{niceblue}{HTML}{8295ff}
\def\bigbox{\color{niceblue}\rule[.25ex]{1ex}{1ex} \hskip .1ex}
\def\squarebullet{\color{black}\rule[0.25ex]{1ex}{1ex}}
\def\singlebox#1{\color{#1}\rule{2ex}{2ex}\color{black} \hskip 1ex}
\def\tolbox#1{\color{#1}\rule[-0.5ex]{2ex}{4.5ex}\color{black} \hskip 1ex}
\def\smallbox{\hskip .25ex \color{niceblue}\rule[.5ex]{.5ex}{.5ex} \hskip .25ex \hskip .1ex}

\def\boxes#1#2{
    \hskip .1ex 
    \newcount\boxnum
    \boxnum=0
    \loop
        \ifnum \boxnum<#1 \bigbox \else \smallbox \fi
        \advance \boxnum by 1
        \ifnum \boxnum<#2
    \repeat
}

\definecolor{grammar}{HTML}{65C2A5}
\definecolor{repetition}{HTML}{FC8D62}
\definecolor{irrelevant}{HTML}{8D9FCB}
\definecolor{contra-sent}{HTML}{E78AC3}
\definecolor{contra-know}{HTML}{A6D853}
\definecolor{common-sense}{HTML}{FFD930}
\definecolor{coreference}{HTML}{E3C494}
\definecolor{generic}{HTML}{B3B3B3}
\definecolor{other}{HTML}{65C2A5}

\definecolor{purple}{HTML}{663399}
\definecolor{light-purple}{HTML}{9370DB}
\definecolor{red}{HTML}{c81d25}

\newcommand{\STAB}[1]{\begin{tabular}{@{}c@{}}#1\end{tabular}}

\maketitle


GPT-Vision has impressed us on a range of vision-language tasks, but it comes with the familiar new challenge: we have little idea of its capabilities and limitations. In our study, we formalize a process that many have instinctively been trying already to develop ``grounded intuition'' of this new model. Inspired by the recent movement away from benchmarking in favor of example-driven qualitative evaluation, we draw upon \textit{grounded theory} and \textit{thematic analysis} in social science and human-computer interaction to establish a rigorous framework for qualitative evaluation in natural language processing. We use our technique to examine alt text generation for scientific figures, finding that GPT-Vision is particularly sensitive to prompting, counterfactual text in images, and relative spatial relationships. Our method and analysis aim to help researchers ramp up their own grounded intuitions of new models while exposing how GPT-Vision can be applied to make information more accessible.

\tableofcontents

\clearpage

\section{Introduction}
\subsection{Motivation}
The recent release of GPT-Vision \citep{openai_gptvision_2023} has prompted widespread excitement---promising to usher in a new era of multi-modal generative AI applications \citep{yang_dawn-lmms-preliminary_2023}. However, a pre-requisite for large-scale utilization of new AI technology is a comprehensive understanding of its associated limitations and failure cases. Without such an understanding, we risk deploying our models in ways that cause real harm to real people---especially in high-stakes domains. 

In this paper we conduct a qualitative example-driven analysis of the various capabilities and limitations of the newly-released GPT-Vision model. Following the lead of \citet{bubeck_sparks-artificial-general_2023}, rather than conducting our analysis in the more traditional way (e.g. collecting a large dataset and computing automatic metrics), we take a more example-driven approach---focusing intently on a small number of illustrative data points and analyzing them extremely closely to glean broader insights and trends. Such an analysis, contrary to the language used in \citet{bubeck_sparks-artificial-general_2023}, has substantial precedent in the social science and human-computer interaction literature and is widely accepted to be scientifically rigorous. 

Furthermore, drawing inspiration from grounded theory \citep{blandford_grounded-theory_2022} and thematic analysis \citep{blandford_thematic-analysis_2022}, we develop and standardize a rigorous method for conducting qualitative analyses of generative AI models. Our method consists of five stages: (1) data collection, (2) data review, (3) theme exploration, (4) theme development, and (5) theme application. As we demonstrate from our findings, such analysis when performed properly allows for deep and intuitive understanding of model capabilities, even when done on relatively small sample sizes.

To illustrate these claims, we focus on one particular task domain: alt text generation for pages and figures in scientific papers. This is a particularly fertile area for analysis, as properly describing the contents of a particular page or figure often requires complex reasoning capabilities that go far beyond simple object detection. Through our analysis we find that GPT-Vision, while extremely impressive, has a tendency to over-rely on textual information, is particularly sensitive to the wording of its prompts, and struggles with reasoning about spatial locality. We are also able to confirm the existence of many of the pitfalls and shortcomings quoted by OpenAI in their model card \citep{openai_gptvision_2023}. Overall, we not only provide insights into the limitations of the newly-released GPT-Vision model but also provide an example of the judicious application of qualitative analysis techniques to generative AI models.

\subsection{Background \& Related Work}
\label{sec:bgrw}
Trying to evaluate the performance of a given model has always been a challenging task. However, recently the rising capabilities of our best models have begun to reveal longstanding shortcomings in our existing evaluations. 

Now that large language models are capable of producing such sophisticated output for a wide range of requests, ``evaluating generated text is now about as hard as generating it'' \citep{neubig_my-nlp-model_2023}. Recent work has warned us against relying on long-used automatic metrics for tasks like machine translation~\citep{fomicheva_taking-mt-evaluation_2019}, question answering~\citep{chen_evaluating-question-answering_2019}, and summarization~\citep{jain_multi-dimensional-evaluation-text_2023, goyal_news-summarization-evaluation_2023} because they may fail to accurately assess novel and abstractive text against a gold standard. Even reference-free metrics have been shown to underestimate the quality of generated text, perhaps because those metrics were trained or evaluated on the same reference-based benchmarks~\citep{goyal_news-summarization-evaluation_2023}. Automatic metrics have long been criticized for unreliably correlating with human judgment, even before the rise of LLMs~\citep{deutsch_statistical-analysis-summarization_2021, belz_comparing-automatic-human_2006}. Reference-free metrics are disproportionately weak at evaluating alt text for blind and low-vision readers~\citep{kreiss_context-matters-image_2022}. Some work has attempted to mitigate these challenges by using an intermediary LLM to evaluate generated text~\citep{liu_g-eval-nlg-evaluation_2023, ding_gpt3-data-annotator_2023a}, designing AI tools to aid data annotation~\citep{gao_collabcoder-gpt-powered-workflow_2023}, or improve metrics and datasets for new LLMs~\citep{jain_multi-dimensional-evaluation-text_2023, zhong_agieval-human-centric-benchmark_2023, sawada_arb-advanced-reasoning_2023}.

Recent example-driven qualitative analyses of GPT-4 and GPT-Vision have already stepped toward robust qualitative analysis for modern LLMs~\citep{bubeck_sparks-artificial-general_2023, openai_gptvision_2023, yang_dawn-lmms-preliminary_2023}. While these studies tend to provide brief commentary on a large number of tasks and examples, we examine a small set of results more deeply through a method based on grounded theory~\citep{blandford_grounded-theory_2022} and thematic analysis~\citep{blandford_thematic-analysis_2022}, which are frequently used in human-computer interaction research. Grounded theory is a data-driven or ``bottom-up'' perspective on data collection and analysis. In grounded theory, patterns and conclusions ``emerge'' from the data, much like an inductive analysis~\citep{bingham_qualitative-analysis-deductive_2023}. Grounded theory instructs analysts to make meaning solely from the data to avoid bias from preconceived notions or existing theories. It includes a method called theoretical sampling, which is based on the idea that we can carefully select data that contains characteristics we care about as opposed to sampling at random or gathering a large dataset~\cite{blandford_sampling_2022}. Theoretical sampling also allows data to be gathered iteratively to address findings as they arise throughout the analysis until we hit ``theoretical saturation'': a subjective yet evidence-based instinct that further data collection and analysis will not reveal any more major insights.

Thematic analysis is a flexible framework through which grounded theory can be applied. First, ``themes'' are gathered from the data, refined, and then applied to the entire dataset to reveal patterns within it. When adopted formally, thematic analysis is a rigorous process that can involve evaluating inter-annotator agreement and setting up infrastructure to protect reliability in qualitative research~\citep{mcdonald_reliability-inter-rater-reliability_2019}. It has been used regularly in well reputed HCI studies like supporting healthcare~\citep{bowman_using-thematic-analysis_2023}, analyzing social media posts~\citep{gauthier_will-not-drink_2022}, and conducting literature reviews~\citep{cooper_systematic-review-thematic_2022}. Brand-new work to be published in the Findings of EMNLP 2023 even proposes an LLM-in-the-loop collaboration framework to assist with thematic analysis~\citep{dai_llm-in-the-loop-leveraging-large_2023}. Our work adapts thematic analysis and grounded theory specifically for evaluating LLMs in NLP research.

Our analysis focuses on GPT-Vision's ability to describe scientific images. Past work on describing images has included automatic image captioning~\citep{tang_vistext_2023, hsu_scicap-generating-captions_2021, spreafico_neural-data-driven-captioning_2020, guinness_caption-crawler-enabling_2018} and alt text generation~\citep{wu_automatic-alt-text-computer-generated_2017, salisbury_scalable-social-alt_2017, williams_quality-alt-text_2022, chintalapati_dataset-alt-texts_2022}. Alt text is a written version of an image that appears in place of it~\citep{vleguru_alt-text-video_2022}. Although alt text is typically associated with screen readers and vision loss, it can also help users with information processing disorders, like issues with visual sequencing, long- or short-term visual memory, visual-spatial understanding, letter or symbol reversal, or color blindness~\citep{mccall_rethinking-alt-text_2022}. Alt text can even help in purely situational circumstances like broken image links or loading issues due to expensive data roaming or weak internet connectivity, which may disproportionately affect individuals with lower incomes~\citep{vleguru_alt-text-video_2022}. Beyond reading online documents, alt text and curated image descriptions can allow audio books and screen readers to ``read aloud'' visual content, giving all of us even more access to news articles, textbooks, blog posts, scientific papers, and other mixed-media texts. These image descriptions, however, need to be generated carefully and, most likely, adaptively.  Alt text by definition depends on the audience, content, and situation, so one approach will not work for all images or people~\citep{w3_images-tutorial_2022}. This claim was validated in practice by a user study~\citep{stangl_going-beyond-one-size_2021}. Blind and sighted readers diverge~\citep{lundgard_accessible-visualization-natural_2021}. Even placement of text has an impact~\citep{stokes_striking-balance-reader_2022}.

Part of this analysis is ``objective,'' such as detecting objects, transcribing labels, and identifying spatial positions, but many aspects are inherently human-centered. What is the ``correct'' interpretation of a graph or the ``main idea'' of a diagram? What is an ``appropriate'' description---not too long, not too short, not too detailed, not too vague? Now that LLMs are so powerful and widely used, we need to address what we as users want from the model beyond just the facts, which we can start to investigate through the human-centered design framework~\citep{norman_design-everyday-things_2013}. We should also acknowledge that different users have different needs, which are often affected by their differing abilities. The same ability can vary in duration and context---consider a user who can needs to have a book read aloud because they are blind, are in a dark room, or had their pupils dilated---as suggested by the ability-based design framework~\citep{wobbrock_ability-based-design-concept_2011}. Together, ability-based human-centered design can help us build inclusive tools for everyone~\citep{Hwang_Build_Your_Own_2023}. 

\subsection{Contributions}

In this paper, we contribute:

\begin{itemize}
    \item Deep, grounded insights on GPT-Vision describing scientific images
    \item A qualitative analysis framework based on grounded theory and thematic analysis for evaluating LLMs
    \item The images we used and the text we generated for future work and reproducibility
\end{itemize}

Part of our goal was to formalize a process that people have already naturally taken to evaluate LLMs: trying a selection of images and prompts, skimming through generated text, and noticing patterns until we are satisfied with our ``intuition.'' Our method provides an organized, systematic framework for intentionally developing this intuition grounded in concrete data. A practical guide on our method and theoretical background will be released soon.
\section{Methods and Data}
\label{sec:methods}

\paragraph{Analysis Procedure} We based our approach to qualitative analysis on well established practices in grounded theory and thematic analysis (see Section \ref{sec:bgrw}). It consisted of five phases: (1) data collection, (2) data review, (3) theme exploration, (4) theme development, and (5) theme application. During the data collection phase, we prompted GPT-Vision to describe a set of scientific figures. We then lightly reviewed the data for notable patterns before carefully searching for ``themes'' during the theme exploration phase. Theme development was dedicated to consulting literature and refining the themes that had emerged in the exploration. Finally, we passed through the data one last time to apply the finalized themes to the entire dataset. This method allowed us to conduct a more rigorous qualitative analysis to gain evidence-grounded intuition about a brand-new model, as we discuss in Section \ref{sec:main-findings}. 

\paragraph{Data collection} During the first phase of our analysis, we collected data through a theoretical sampling approach \citep{blandford_sampling_2022}. We were initially interested in GPT-Vision's ability to describe scientific figures and eventually expanded to images of code, math, and even full pages from research publications. Our final set of images contained two photos, three diagrams, four graphs, three tables, five screenshots of full pages, three images with computer code, and two images with mathematical notation for a total of 21 images \mbox{(see Appendix Tables \ref{tab:data_figures} and \ref{tab:data_full-special})}. For figures, we included the texts of the caption and a reference paragraph as context as well. We queried GPT-Vision with the following two prompts for each image, giving us a total of 42 generated passages:

\begin{quote}
    ``alt'': Write alt text for this $<$input$>$ .

    ``desc'': Describe this $<$input$>$ as though you are speaking with someone who cannot see it.
\end{quote}

\noindent We replaced $<$input$>$ with ``figure'' for photos, diagrams, and graphs; ``table'' for tables; ``page'' for screenshots of full pages; and ``image'' for images of special text (code or math).\footnote{Images, context, and generated passages can be found at \url{https://github.com/ahwang16/grounded-intuition-gpt-vision}.}

\paragraph{Data review} After settling on a preliminary set of scientific images, we generated passages with GPT-Vision and skimmed them for prominent patterns and surprises. We recorded these initial observations in ``memos,'' a flexible form of taking notes \mbox{\citep{blandford_thematic-analysis_2022}}. The goal of this process was to gain familiarity with our data as a whole in preparation for theme exploration. We periodically noticed some trends that we wished to investigate further during this phase. Following the theoretical sampling methodology, we prompted GPT-Vision for more data as insights surfaced from our initial image set \citep{blandford_sampling_2022}. Additional images for ``one-off experiments'' are included in Appendix Table \ref{tab:data_figures} as P1.1 and T1.1.

\paragraph{Theme exploration} Usually called ``open coding'' or ``open pass'' in grounded theory methodology, this phase focused on discovering patterns---typically termed ``themes'' or ``codes''---within the data \citep{corbin_grounded-theory-research_1990}. We carefully read each generated passage, recording themes and evidence (e.g., quotes) in a structured document. Diverging from original methodology, we consulted relevant literature to inform the final themes. We also conducted ``aggregate analyses,'' a new step we established specifically for evaluating generative AI models. In traditional approaches, data is inspected one at a time, with insights from the pool of previous data guiding the next analysis. In an aggregate analysis, we directly compared groups of related passages (e.g., all graphs). At the end of our exploration, we had a hierarchy of 51 preliminary themes like hallucination, numerical reasoning, writing style, and contextual influence.

\paragraph{Theme development} We finalized our themes during the theme development phase by renaming, redefining, removing, creating, merging, or splitting themes from the exploration phase as needed. This phase was based on ``axial coding'' from the original grounded theory methodology, in which themes are grouped together if they share a connection of ``axis'' of similarity \cite{corbin_grounded-theory-research_1990}. At the end of the development phase, our finalized hierarchy consisted of 94 themes. We present a sample of these themes in Table~\ref{tab:theme_examples}.\footnote{The full set of themes can be found at \url{https://github.com/ahwang16/grounded-intuition-gpt-vision}.}

\begin{table}[]
\renewcommand{\arraystretch}{1.2}
\small
\centering
\begin{tabular}{ll}
\toprule
\multicolumn{1}{c}{Theme} & \multicolumn{1}{c}{Definition} \tabularnewline
\midrule
Linguistic characteristics & General features of text generated by GPT-Vision \tabularnewline
\hspace{2mm}  \textbullet \hskip 1.0ex Persona & GPT-Vision's ``personality,'' attitude, or tone of voice \tabularnewline
\hspace{6mm} \squarebullet \hskip 1.0ex Stoic authority & Matter-of-fact, assertive, straightforward (aloof and certain) \tabularnewline
\hspace{6mm}  \squarebullet \hskip 1.0ex Customer service rep & Conversational, polite, easygoing (engaging and uncertain) \tabularnewline
\hspace{2mm} \textbullet \hskip 1.0ex First-person language & Instances of first-person language (I/me/my/mine, we/us/our/ours) \tabularnewline
Figure descriptions & Characteristics of how standalone elements with a caption are described \tabularnewline
\hspace{2mm} \textbullet \hskip 1.0ex Main idea & The purpose or critical message of a figure, if described \tabularnewline
\bottomrule
\end{tabular}
\caption{Examples of finalized themes after theme development. Indentations represent sub-themes that were categorized under a larger parent theme (e.g., ``Stoic Authority'' is a sub-theme of ``Persona,'' which is a sub-theme of ``Linguistic characteristics'').}
\label{tab:theme_examples}
\end{table}

\paragraph{Theme application} During the final phase of our analysis, we passed through the data another time to apply our finalized themes. For each generated passage, we recorded any overlooked evidence that fit into a theme. The outcome of this phase was a detailed document of themes and evidence. These themes were the building blocks for our findings (see Section \ref{sec:main-findings}), analogous to ``latent representations'' for our ultimate conclusions.
\section{Findings}
\label{sec:main-findings}

In this section, we discuss our findings across all images and prompts. We refer to images with a letter signifying the image type (\textbf{P}hoto, \textbf{D}iagram, \textbf{G}raph, \textbf{T}able, \textbf{C}ode, \textbf{M}ath, or \textbf{F}ull page) and a number. A list of all images can be found in Tables \ref{tab:data_figures} (photos, diagrams, graphs, and tables) and \ref{tab:data_full-special} (full pages, code, and math). We used two prompts for each image (see Section \ref{sec:methods}). The first prompt, which we call ``alt,'' is a straightforward request for alt text. The second prompt is identified as ``desc'' and instructs GPT-Vision to describe the image as though it were speaking with someone who could not see it. The ``alt'' prompt often resulted in a paragraph with a matter-of-fact tone, while most generated passages for the ``desc'' prompt were about a page long with varying levels of cheerfulness. All images and generated passages can be found at \url{https://github.com/ahwang16/grounded-intuition-gpt-vision}.

\subsection{Margin for Error}
\label{sec:margin-for-error}

One of the most apparent patterns of GPT-Vision's writing style was how much margin for error it included in the generated passage. As mentioned in its system card, GPT-Vision can sometimes speak in a matter-of-fact tone \citep{openai_gptvision_2023}. Other times, it implies imprecision---this \textit{or} that, it \textit{seems} like, and so on.

\begin{figure}[h]
    \centering
    \includegraphics[width=0.4\textwidth]{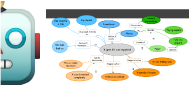}
    \caption{Robot emoji and knowledge graph from D3 \citep{west_symbolic-knowledge-distillation_2022}.}
    \label{fig:d3_distillation}
\end{figure}

\paragraph{Sometimes counterbalancing an error} Sometimes, a wide margin for error compensated for a mistake GPT-Vision made, like describing elements in a complicated diagram. D3, the front-page figure representing symbolic knowledge distillation, is particularly complex. Symbolic knowledge distillation involves training a language model to generate commonsense knowledge graphs, which are then used to train other commonsense models \citep{west_symbolic-knowledge-distillation_2022}. GPT-Vision described two parts of D3 with notable margin for error: the standard Apple of a robot's face and a connected graph of nodes and edges, shown below (Figure \ref{fig:d3_distillation}).

In D3, the robot emoji is used multiple times to represent different language models. One of the robots shows only half of its face, revealing one large eye and a red ``ear.'' GPT-Vision described it as

\begin{quote}
    a cute character with a square-shaped body, has a single large eye, and a small red part on its side that I assume represents an arm or [sic] sorts. (D3 desc)
\end{quote}

\noindent GPT-Vision made a couple errors: it described the half-emoji as having a ``square-shaped body'' even though it has just a face. It exhibited some margin for error by saying ``\textit{I assume}'' a small red part is an arm, which is incorrect. Appropriate margin for error can help the user develop trust in the model, as long as the implied uncertainty matches the actual accuracy of the claim.

P1 desc, T1 desc, T2 desc, F4 desc, D1 desc, G1 alt and desc, G2 desc, and C2 desc contain similar behavior.

\paragraph{Occasionally distracting or excessive} While useful for indicating uncertainty, margin for error occasionally cluttered GPT-Vision's output with too much bloat. For example, when describing the connected graph of nodes and edges, it wrote

\begin{quote}
    there is an illustration of what appears to be some form of inter-connected network, which may be meant to visually represent the structure of a knowledge graph. (D3 desc)
\end{quote}

\noindent Ironically, GPT-Vision sometimes sounded confident when it was wrong and unsure when it was right. This level of hedging may especially confuse readers without access to the image because they cannot interpret it for themselves.

C3 desc and M2 desc contain similar behavior.







\paragraph{Often necessary} Including some margin for error was often necessary because GPT-Vision's claim could not be verified by the input alone. Even a detail as seemingly obvious as

\begin{quote}
    The image is a photo of a section of an academic paper or textbook, focused on a specific topic titled ``3.1 Decoder: General Description.'' (M1 desc)
\end{quote}

\noindent cannot be confirmed with the input GPT-Vision has been given. It correctly guessed the origin of M1---a section from \citet{bahdanau_seq2seq-attention_2016}---but the image itself does not state that it is from an academic paper. Depending on how confident we are in GPT-Vision's internal knowledge, tuning the margin for error empowers users to make informed decisions with LLM assistance.



C1 desc, C3 desc, P1 alt and desc, F1 desc, F2 desc, F3 desc, F4 desc, T3 desc, M1 desc, M2 desc, and D1 desc contain similar behavior.

\subsection{Hallucination}
\label{sec:hallucination}

Hallucination was one of the main vulnerabilities listed in GPT-Vision's system card, but we argue that not all hallucination needs to be avoided. In fact, some forms of hallucination are highly desired.

\paragraph{Hallucination as general knowledge and inference} When defining it as information in the output that does not appear in the input, then general knowledge and inference can be considered helpful forms of hallucination.

GPT-Vision displayed several signs of ``internal knowledge'':

\begin{itemize}
    \item ``Egg Biryani is an Indian dish'' (P1 desc).
    \item ``The page has mathematical symbols and technical terms commonly found in computer science literature'' (F5 alt).
    \item ``[The Python code] uses comments (text preceded by a `\#' symbol'') (C3 desc).
\end{itemize}

\noindent many of which are accurate. P1 desc, D2 desc, C2 alt and desc, and M2 desc contain similar behavior.

GPT-Vision made reasonable inferences even more often. These claims seemed reasonable given the input but are not stated directly within it, like

\begin{itemize}
    \item ``...another gray dashed horizontal line near the top, labeled `Human', [indicates] the human-level performance benchmark'' (G3 alt).
    \item ``[$\alpha_{xy}$] probably refers to a certain value that depends on x and y'' (C1 desc).
\end{itemize}

\noindent One perspective on natural language inference relates to the model's ability to reason. In our case, we see it as a hallucination that happens to be correct.

C1 desc, C2 desc, D1 desc, D2 desc, D3 desc, F2 desc, G3 desc, M1 desc, and M2 desc contain similar behavior.

\paragraph{Beware of possibly naive assumptions} A handful of ``interpretations'' seem like valid and impressive inferences, but we cannot know for sure without additional studies on internal model mechanisms. In a particularly subtle but impactful instance, GPT-Vision described the trend of the table of errors in T2 as

\begin{figure}[h]
    \centering
    \includegraphics[width=0.7\textwidth]{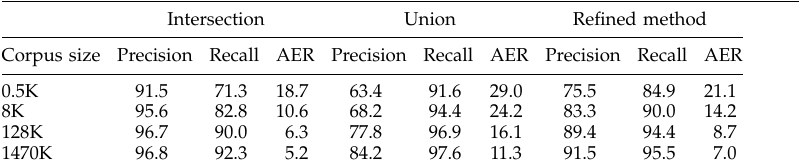}
    \caption{A table of performance metrics from \citet{bahdanau_seq2seq-attention_2016} (T2).}
    \label{fig:t2_quantitative}
\end{figure}

\begin{quote}
    The values in these columns generally decrease as the size of the training corpus increases, indicating improved performance with more data. (T2 alt)
\end{quote}

\noindent At first glance, this claim seems reasonable---impressive, even. We may be surprised that decreasing values indicate ``improved performance,'' but even this makes sense knowing that the values are error rates. However, we should be careful before assuming that GPT-Vision can ``read'' tables. This assertion may have been a lucky coincidence because model performance often improves in general as the training corpus increases. Our analysis of ``artificial behavior'' focuses on capturing these external patterns, which should not be conflated for the underlying processes of ``artificial cognition'' or the mechanical structures of ``artificial neuroscience.''

F3 alt, F1 alt, T3 alt, F5 alt, and C3 desc contain similar behavior.

\subsection{Incorporation of Source Material}
\label{sec:inc-source-material}

Conversely from hallucinating, GPT-Vision also incorporated source material in a few ways.

\paragraph{Direct quotes} GPT-Vision commonly provided exact section headers, text in diagrams, and publication metadata as direct quotes. Some of these quotes helped describe the structure of the image:

\begin{quote}
    The left column then lists ``CSS CONCEPTS'', which look like categories that the article might belong to, and is followed by one entry that reads `` $\bullet$ Human-centered computing $\rightarrow$ Interactive systems and tools.'' (F1 desc)
\end{quote}

\noindent GPT-Vision replicated section exactly, down to the bullet point. Future models in human-centered applications can consider elaborating special formatting even more, especially if the content will played by audio books and screen readers.

\begin{figure}[h]
    \centering
    \includegraphics[width=0.3\textwidth]{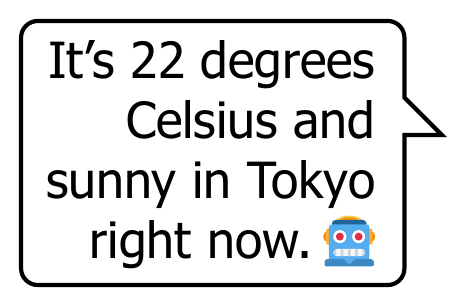}
    \caption{The chatbot's response from D1 \citep{zhu_kani_2023}.}
    \label{fig:d1_kani_response}
\end{figure}

Exact direct quotes are also used to indicate specific words from the image:

\begin{quote}
    [The chatbot] says, ``It's 22 degrees Celsius and sunny in Tokyo right now.'' (D1 desc)
\end{quote}

\noindent Overall, GPT-Vision displayed impressive ability to recognize text in images, which will be a great strength for describing images in general. As noted in \citep{openai_gptvision_2023}, it sometimes makes errors, especially when two text elements are close to each other. It understandably made more mistakes on smaller or blurrier text, which it could indicate to the user to help them judge the quality of GPT-Vision's descriptions.

C1 desc, P1 desc, T1 desc, F2 alt, T3 alt, F4 alt, D1 desc, D2 desc, G2 desc, G3 alt and desc, and G4 desc demonstrate similar behavior.

\paragraph{Slightly altered quotes} Some of the direct quotes were slightly different from the original text, which may misrepresent the intent of the original author. These altered quotes were occasionally benign, like removing the hyphens in ``Herb-Roasted Salmon with Tomato-Avocado Salsa'' (F1 alt) or capitalizing ``method'' in ``The columns from left to right are titled `Corpus size', `Intersection', `Union', and `Refined Method'{}'' (T2 desc).

Other changes were more severe, however. The last name of one of the authors of F5, Frohlich, was misprinted as ``Fritsche'' even though the other three names were spelled correctly (alt).

In addition, GPT-Vision sometimes omitted parts of the text that affected its meaning, such as removing ``search'' from a figure title: 

\begin{quote}
   ...there is a figure titled ``Fig. 1. Generators for binary \sout{search} trees.'' (F5 alt) 
\end{quote}

\noindent The lead author of this paper confirmed that this modification misrepresents the caption because not all binary trees are binary \textit{search} trees and binary search trees in particular were important for that figure.

We also witnessed one instance of GPT-Vision merging nearby text elements in a figure, which was mentioned in \citet{openai_gptvision_2023} (D2 alt) When quoting the original source, LLMs should represent the source accurately or indicate where changes were made with brackets, ellipses, or other devices.

\paragraph{``Plagiarism''} 

GPT-Vision often generated text that was very similar to the source. In the following example, the \textbf{boldface} text from the generated passages appears verbatim in the source:

\begin{quote}
    This text explains that the goal of the study is \textbf{to understand how voice assistants can effectively guide people through complex tasks,} like following \textbf{recipes}. (F2 alt)
\end{quote}

\noindent Comparing this passage to the original text makes it sound eerily familiar:

\begin{quote}
    We designed an observational study \textbf{to understand how voice assistants can effectively guide people through complex tasks,} using \textbf{recipes} as an example. \citet{hwang_rewriting_2023}
\end{quote}

Some examples seem ``paraphrased'' (\uwave{emphasis} ours)

\begin{quote}
    \uwave{The context vector} is \uwave{computed} by an RNN and \uwave{relies} on a \uwave{sequence of annotations}, with \uwave{each annotation containing information about the whole input sequence with a focus on surrounding parts of a specific word}. M1 alt
\end{quote}

\noindent but still too similar to the original text to be acceptable (bracketed ellipsis [...] ours).

\begin{quote}
    \uwave{The context vector} $c_i$ depends on a \uwave{sequence of \textit{annotations}} $(h_1, \ldots, h_{T_z})$ [...] \uwave{Each annotation $h_i$ contains information about the whole input sequence with a strong focus on the parts surrounding the $i$-th word} of the input sequence. [...] The context vector $c_i$ is, then, \uwave{computed} as a weighted sum [...] \citep{bahdanau_seq2seq-attention_2016}
\end{quote}

In the worst case scenario, these kinds of reproduction could be flagged as plagiarism or copyright violation. LLMs have strong potential to help with complex writing tasks from crafting emails to redrafting reports, so they should be carefully tuned to quote significant amounts of reproduced text and paraphrase properly. 

\subsection{Sensitivity to Typographical Influence}
\label{sec:typographical-influence}

Sometimes, leaning too much on text in an image for context is risky. GPT-Vision was particularly prone to typographical attacks, reminiscent of its predecessor CLIP \citep{goh_multimodal-neurons-artificial_2021} and related to the vulnerability to the order of images mentioned in GPT-Vision's system card \citep{openai_gptvision_2023}.

\begin{figure}[h]
    \centering
    \includegraphics[width=0.8\textwidth]{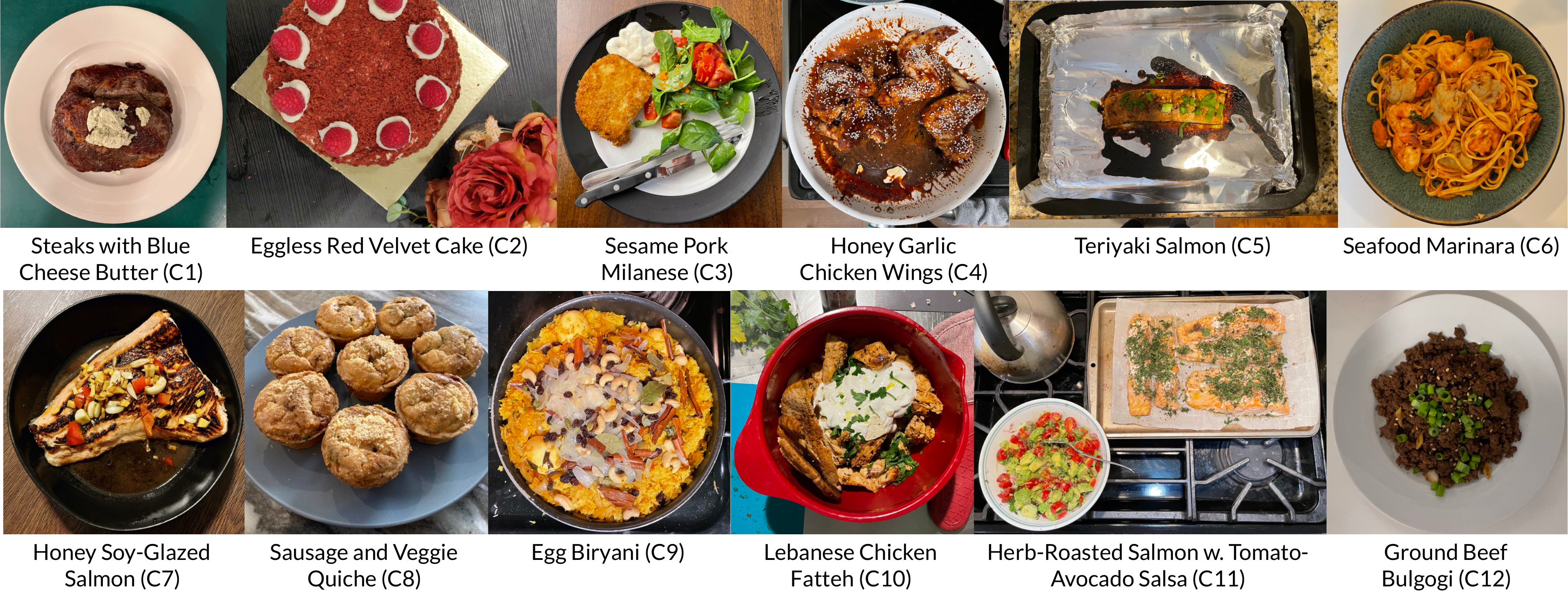}
    \caption{Dishes prepared by participants in a recent study (P1) \citep{hwang_rewriting_2023}.}
    \label{fig:p1_food}
\end{figure}

\paragraph{Successfully incorporating original labels} One of our images, P1, consists of a 2x6 grid of photos showing twelve dishes prepared by participants in a recent study \citep{hwang_rewriting_2023}. The photos are also labeled with the participant's identification number and the name of the dish underneath each one.

The first photo shows ``(C1) Steaks with Blue Cheese Butter,'' which GPT-Vision aptly described as 

\begin{quote}
    (C1) A perfectly cooked steak topped with blue cheese butter on a white plate. (P1 alt)
\end{quote}

\noindent All of the dishes in this passage incorporate the corresponding label in some way, and nearly all of them are excellent, suggesting that text in images can be a helpful source of context.

\begin{figure}[h]
    \centering
    \includegraphics[width=0.8\textwidth]{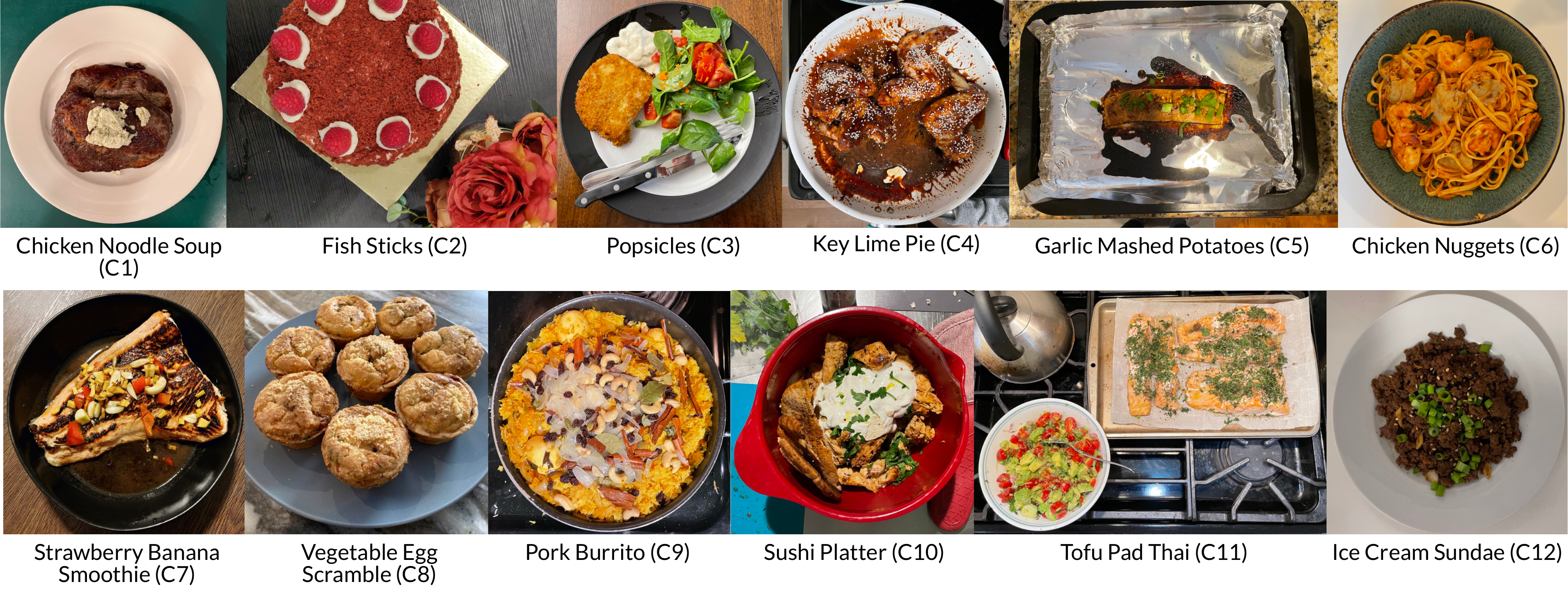}
    \caption{The same figure as P1, but with adversarial labels (P1.1) \citep{hwang_rewriting_2023}.}
    \label{fig:p1_food_modified}
\end{figure}

However, in a one-off experiment with adversarially modified labels, this blessing turned into a curse. We labeled the same dish as ``Chicken Noodle Soup,'' which GPT-Vision continued to incorporate:

\begin{quote}
    (C1) Chicken Noodle Soup, where a bowl is presented with a dark broth and a dollop of cream... (P1.1 alt)
\end{quote}

\noindent The photo clearly shows dark, cooked meat topped with a scoop of butter-like dressing, but GPT-Vision still tried to incorporate the new label. All twelve photos were similarly affected in both the ``alt'' and ``desc'' generated passages (P1.1).

\paragraph{Reality check} When asked to verify if the given labels were correct and provide alternatives otherwise, GPT-Vision's descriptions improved for both the original 

\begin{quote}
    ...The label is correct. The photo shows a steak with a pat of blue cheese butter on top.    
\end{quote}

\noindent and adversarial labels.

\begin{quote}
    The label is incorrect. The photo shows what appears to be a steak with butter on top. The correct label could be ``Steak with Butter''.
\end{quote}

These corrections sometimes sound very certain (see Section \ref{sec:margin-for-error}), leading GPT-Vision to provide some inaccurate descriptions in an authoritative tone. This is a known limitation specified in its system card \citep{openai_gptvision_2023}.

\paragraph{Hazarding a guess} GPT-Vision also performed moderately well when presented the photos without labels (P1.2).

\begin{quote}
    ...a seared steak with butter... (P1.2 alt)
\end{quote}

GPT-Vision is clearly a powerful vision model, and it can become even more powerful by learning to mitigate the extent to which text in an input image can change the way GPT-Vision talks about it.

\subsection{Lossy Expansion}
\label{sec:lossy-exp}

During our investigation, we conducted a one-off experiment to evaluate GPT-Vision's performance on figures when they are contained within larger images. When describing the two-by-six grid of food photos in P1, the ``alt'' passage correctly stated there were 12 photos (although it incorrectly characterized the figure as a ``4x3 grid''). The ``desc'' did not specify a total number of photos, but it correctly stated that the figure was ``in a two-row rectangular format with six dishes displayed on each row.''

\begin{figure}[h]
    \centering
    \includegraphics[width=0.7\textwidth]{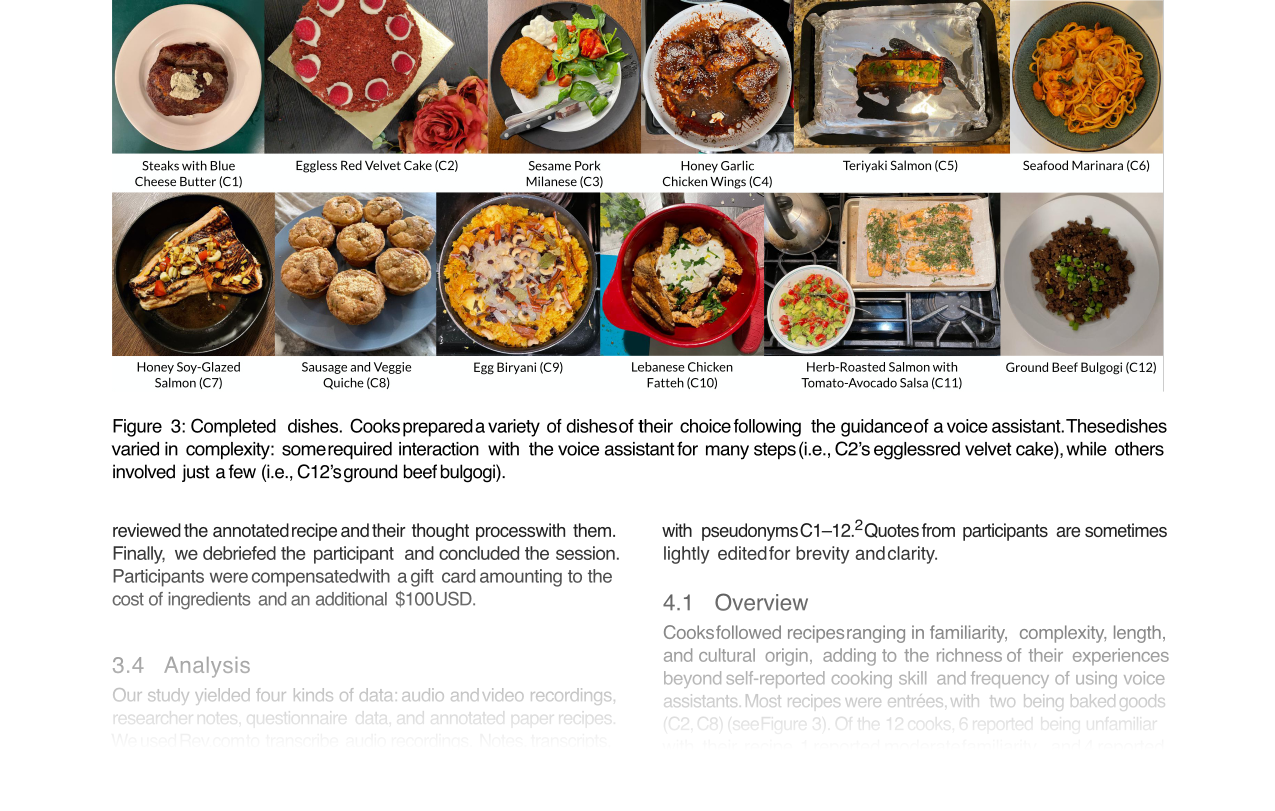}
    \caption{The beginning of F2, a page from \citet{hwang_rewriting_2023}.}
    \label{fig:f2_photos_counting}
\end{figure}

We were expecting to see similar behavior for F2, which is a screenshot of the page that contains P1, but GPT-Vision instead claimed that there were 10 (alt) and 8 (desc) photos in the grid. We attempted to investigate further, but asking GPT-Vision to start analyzing F2 by focusing on the figure first did not improve the results. Asking it directly for the number of photos did not help either. This could cause problems down the line when users are asking for descriptions of arbitrary images. GPT-Vision may err on subcomponents of an image and users may not think to provide the subcomponent on its own and try again, especially if GPT-Vision is providing an accessibility service for an image they cannot see.

\begin{figure}[h]
    \centering
    \includegraphics[width=0.7\textwidth]{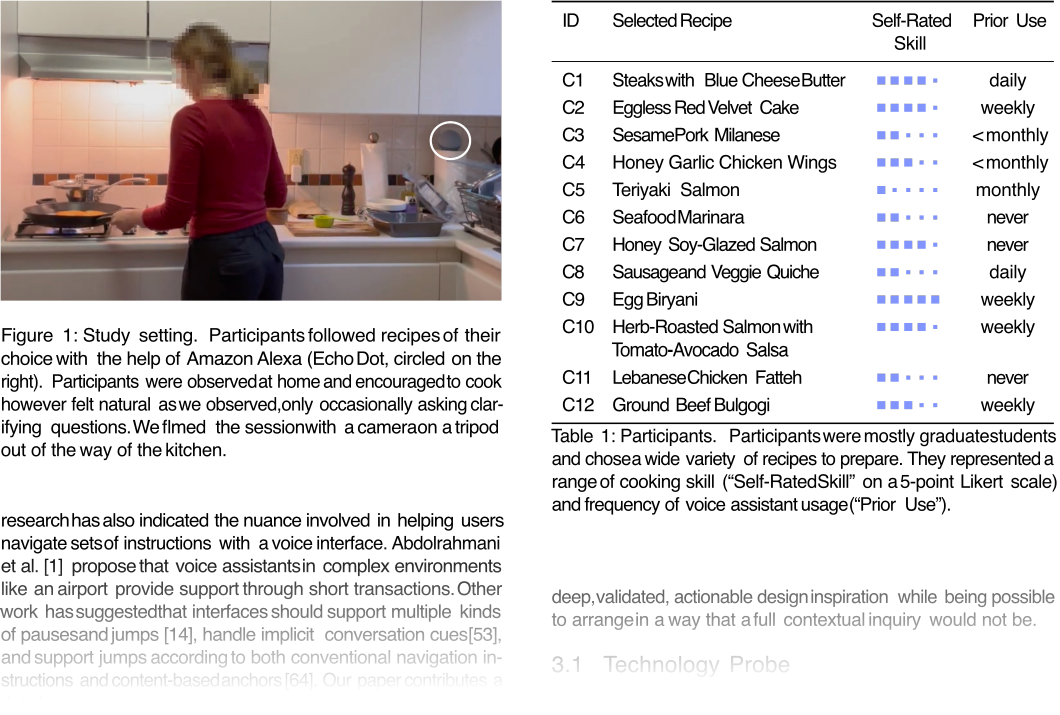}
    \caption{An excerpt from F3, the third page of \citet{hwang_rewriting_2023}.}
    \label{fig:f3_photos_counting}
\end{figure}

We noticed similar behavior with the table in T1 and the full page it appears in (F2). T1 is a table from the same study that contains a list of recipe titles matching the names of the dishes in P1. GPT-Vision showed no issues reproducing the recipe titles in T1, but it suddenly started making errors when listing the same recipe titles from the same table in F2. ``Eggless Red Velvet Cake'' turned into ``Eggs Red Velvet Cake'' (alt) or ``Egg Fried Rice'' (desc) and ``Sesame Pork Milanese'' became ``Sesame Pork Medallions,'' among other new errors.

Our analysis of ``artificial behavior'' focuses on exposing these patterns rather than inferring why they occur, so we are unsure why GPT-Vision sometimes describes the same elements drastically differently---if this is a frequent problem at all. Our small sample size of 21 images may not have much statistical power, but a phenomenon occurring more than once is bound to be compelling in such few cases.

In fact, P1 and T1 were two of only three images that also appeared in full-page screenshots, constraining our sample even more. The third image, P2 could be an outlier: unlike P1 and T1, it does not contain any text. We speculate that the way images are ``tokenized'' may lead to these errors, and further investigation into the ``artificial cognition'' and ``neuroscience'' of these behaviors will hopefully reveal the answer.

\subsection{Taking Context into Consideration}
\label{sec:context}

\begin{figure}
    \centering
    \includegraphics[width=0.5\textwidth]{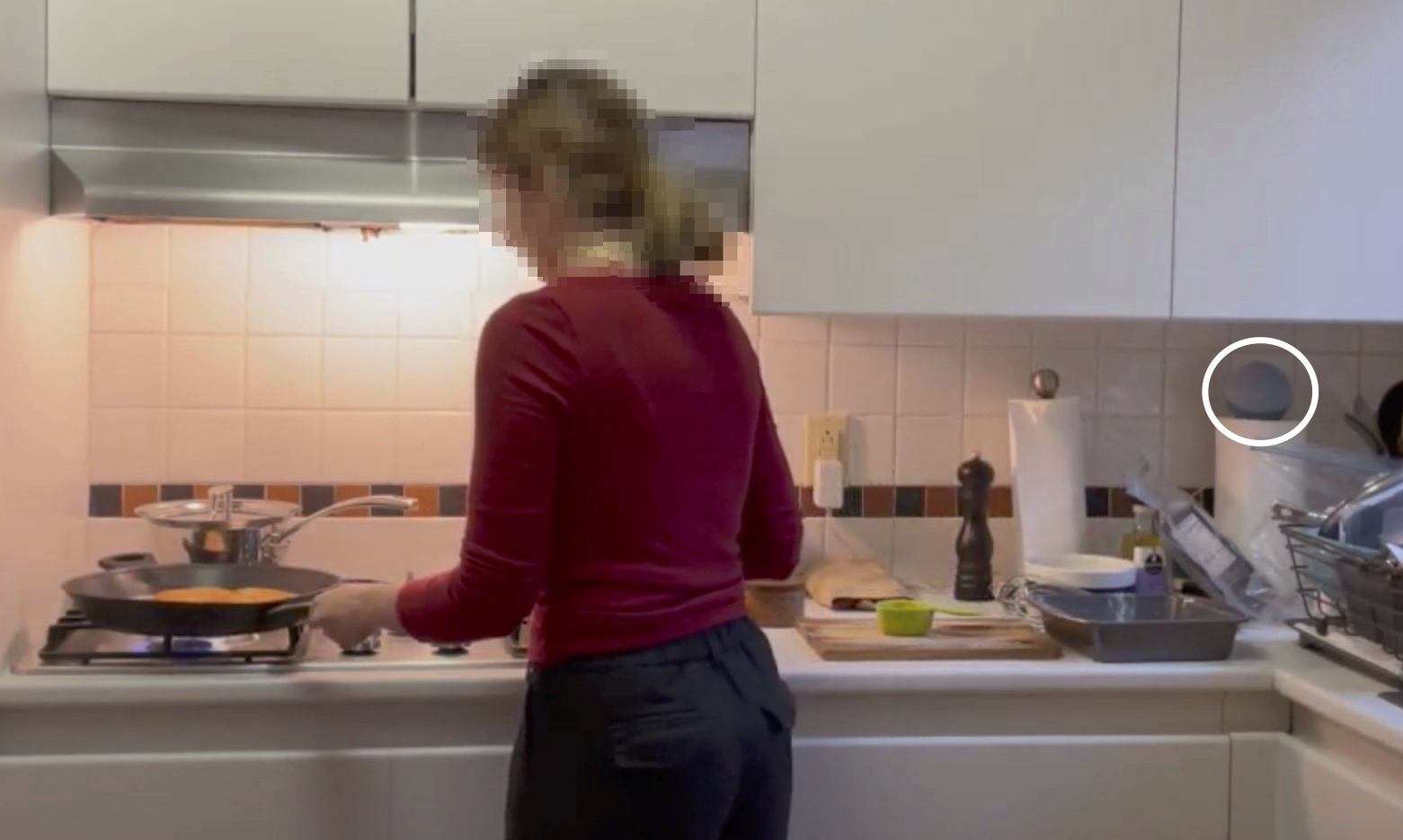}
    \caption{A photo of a participant cooking in a kitchen with an Amazon Alexa Echo Dot circled on the right (P2) \citep{hwang_rewriting_2023}.}
    \label{fig:p2}
\end{figure}

P2 is a photo of a female study participant cooking a dish in a kitchen with the guidance of a voice assistant. The voice assistant, a blue spherical Alexa Echo Dot, is displayed on the right side of the image with a circle around it.

Although the type of device circled in image is not immediately clear, the caption that was given as additional context stated that ``[participants] followed recipes of their own choice with the help of Amazon Alexa (Echo Dot, dircled on the right)'' \citep{hwang_rewriting_2023}.

\paragraph{Taking a hint} GPT-Vision incorporated this information well for the ``desc'' prompt,

\begin{quote}
    What stands out is an Amazon Alexa Echo Dot, which is circled for emphasis. It is placed to the far right on the countertop near some other kitchen tools. (P2 desc)\end{quote}

\paragraph{Missing the point} but not for the ``alt'' prompt.

\begin{quote}
    There is a small circular clock with a white frame hanging on the wall, indicated by a circle. (P2 alt)\end{quote}

In most cases, information from the context did not appear in the generated passages. This suggests that GPT-Vision ``ignored'' it, but we cannot know for sure based on our behavioral evaluation. We do, however, have evidence that GPT-Vision has the capacity to leverage text and image inputs at the same time.

D3 alt, P1 desc, P2 desc, and G2 alt show similar examples of incorporating context. D3 alt and desc, T1 alt and desc, P2 alt, T2 alt, T3 alt and desc, D1 alt and desc, G1 alt and desc, G2 alt and desc, and G4 alt and desc contain similar behavior of lacking context.

\subsection{Code-to-English Translation}
\label{sec:code}

In general, GPT-Vision's descriptions of code demonstrated some internalized knowledge of programming languages at a high-level. The specificity of the Python description compared to Haskell suggests a deeper knowledge of Python, while its fluent ``translation'' of pseudocode to natural language indicates good potential in the programming space.

\begin{figure}[h]
    \centering
    \includegraphics[width=0.75\textwidth]{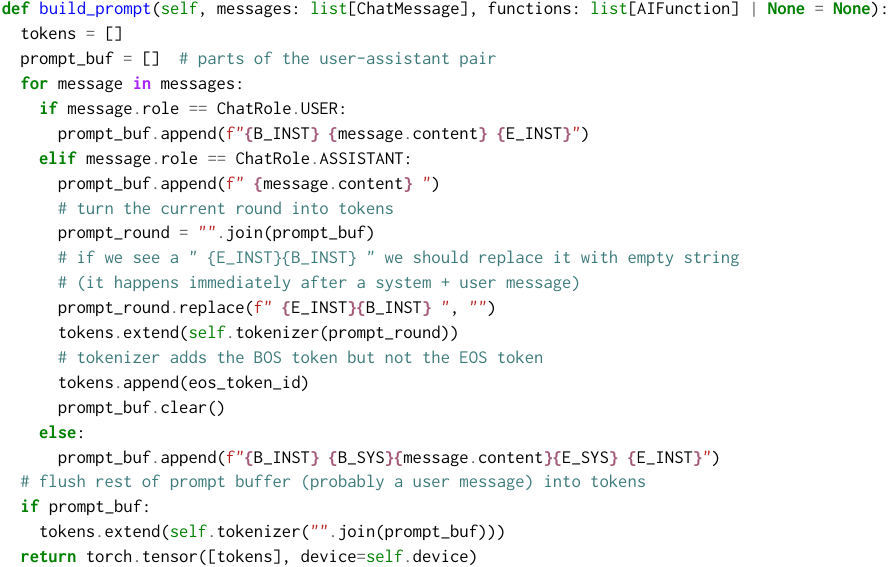}
    \caption{The \texttt{build\_prompt()} method from C3 \citep{zhu_kani_2023}.}
    \label{fig:c3_python_build-prompt}
\end{figure}

\paragraph{Python} GPT-Vision correctly indicates that C3 ``contains a screenshot of Python code which defines a class named `LlamaEngine' that inherits from `HuggingEngine'{}'' (C3 alt, and that the class ``has three methods: `\_\_init\_\_', `build\_prompt', and `message\_len'{}'' (C3 alt)

It even elaborates on the methods, such as correctly stating that ``the `build\_prompt' is meant for building and tokenizing a prompt from a user-assistant conversation, using incoming messages and functions. It accepts messages and functions as parameters and appends tokens to build a prompt'{}''  (C3 alt)

When queried with the ``desc'' prompt, GPT-Vision provides even more specific details: 

\begin{quote}
    the \`{}build\_prompt\`{} method... accepts two parameters: `self', which is standard for class methods, and 'messages', which is expected to be a list of ChatMessage. (C3 desc)\end{quote}

\noindent These details, while correct, may not be the most helpful overview of a piece of such a sophisticated code. It incorporates special tokens depending on the chat role and specifies particular Python type annotations. GPT-Vision even misprints one of the types as ``SomeFunction'' rather than ``AIFunction'' (C3 desc). Generative AI models describing code should look for the most critical structures within it, which may not be the most obvious pieces.

\begin{figure}[h]
    \centering
    \includegraphics[width=0.75\textwidth]{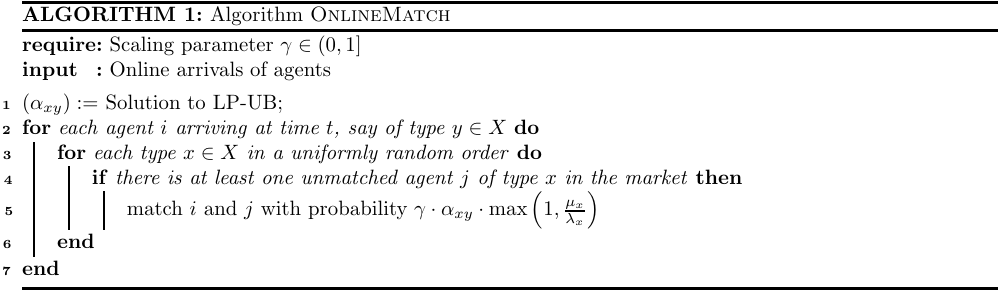}
    \caption{A pseudocode algorithm for a dynamic matching (C1) \citep{collina_algorithm-dynamic-weighted-matching_2021}.}
    \label{fig:c1_pseudocode}
\end{figure}

\paragraph{Pseudocode} GPT-Vision also ``translates'' pseudocode to natural language quite well, besides some errors in reproducing mathematical text (see Section~\ref{sec:math} for more details). C1 contains an if statement in two nested for loops to demonstrate an dynamic matching algorithm. Instead of reproducing the pseudocode verbatim, it describes it more generally:

\begin{quote}
    The algorithm starts by calculating the solution to `LP-UB' and stores it in `$\alpha xy$'. Then, for each agent `i' arriving at time `t' of a certain type `y', and for each type `x' in a random order, it check if there's at least one unmatched agent `j' of type `x'. If so, agent `i' is matched with `j' with a calculated probability dependent on `$\gamma$' and `$\alpha xy$'. The algorithm terminates after processing all agents. (C1 alt)
\end{quote}

\noindent When responding to the ``desc'' prompt, it breaks the pseudocode down into four steps:

\begin{enumerate}
    \item First, it calculates ``(a\_xy)''...
    \item The algorithm then enters a loop where for each agent ``i'' ..., it further loops through each type ``x''...
    \item Inside this nested loop, there is a conditional statement...
    \item If this condition is true, the algorithm matches agent ``i'' and ``j''... (C1 desc)
\end{enumerate}

\paragraph{Haskell} Descriptions of Haskell, however, tend to be much more superficial, like stating that C2 starts with ``the definition for a data type called \`{}Freer\`{}, followed by definitions for \`{}{}Return\`{}{} and \`{}{}Bind\`{}{}'' (desc, or ``[the] type alias \`{}Reflective\`{} [is] defined as \`{}Freer (R b)\`{}'' (alt) The difference between GPT-Vision's behavior with Python and Haskell may suggest that it is ``more familiar'' with the former.\textbf{}

\begin{figure}[h]
    \centering
    \includegraphics[width=0.7\textwidth]{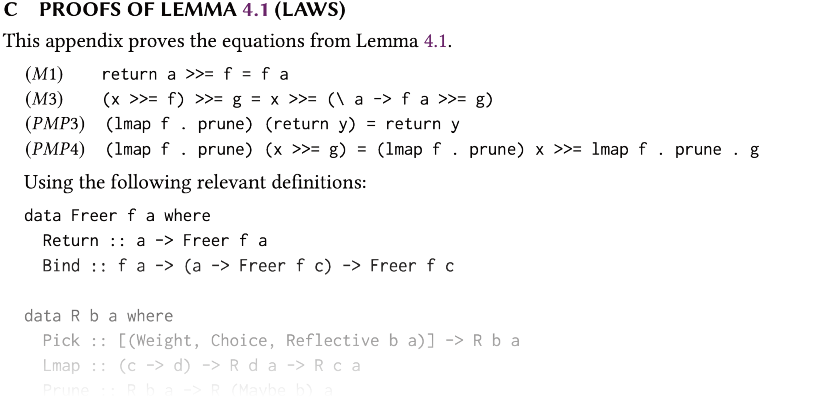}
    \caption{The beginning of a Haskell proof (C2) \citep{goldstein_pl-reflecting-random_2023}.}
    \label{fig:c2_haskell_translation}
\end{figure}

\subsection{Visions of Summarization}
\label{sec:summarization}

To our surprise, when given screenshots of full images, GPT-Vision often showed signs of summarizing paragraphs within them. The ability for vision models to handle sophisticated language tasks like summarization opens many opportunities for them to handle dense, text-dominant documents as well as the images within them.

One full-page screenshot (F3) that GPT-Vision started to summarize was the third page of \citet{hwang_rewriting_2023}, which discusses a human-computer interaction study on how voice assistants tend to deliver complex instructions (see Figure \ref{fig:f3_summary}). It looked like a coherent summary at first glance, but a deeper look showed us that most of it was composed of paraphrased sentences from throughout the document. After detailing the layout of the two-column paper, GPT-Vision described portions of the page starting with the left column:

\begin{figure}[h]
    \centering
    \includegraphics[width=0.7\textwidth]{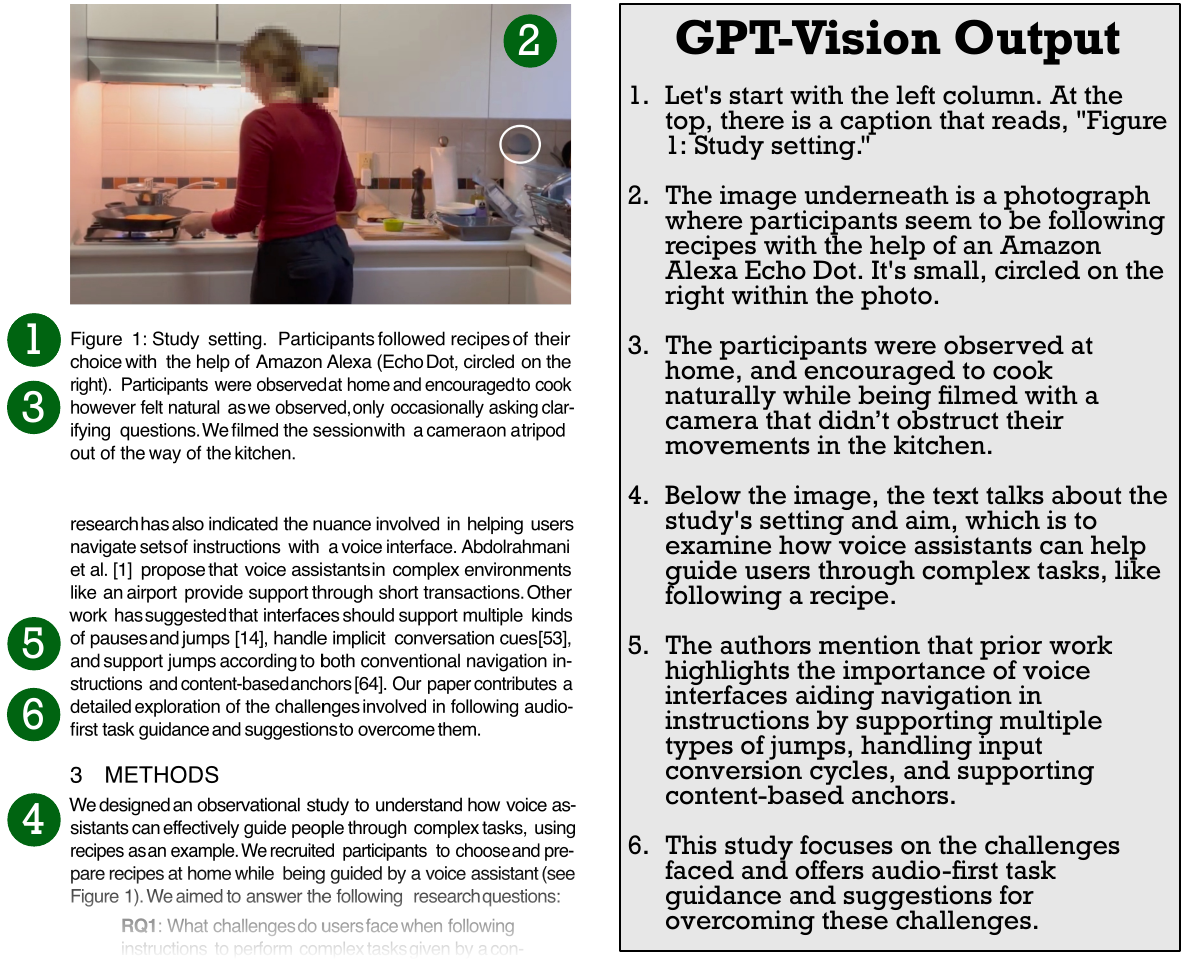}
    \caption{Part of the left column of F3 and a portion of GPT-Vision's description of it \citep{hwang_rewriting_2023}.}
    \label{fig:f3_summary}
\end{figure}

\begin{quote}
    Let's start with the left column. At the top, there is a caption that reads, ``Figure 1: Study setting.'' The image underneath is a photograph where participants seem to be following recipes with the help of an Amazon Alexa Echo Dot. It's small, circled on the right within the photo.
\end{quote}

This part of the summary quoted and paraphrased the caption of Figure 1, which is at the top of the left column. GPT-Vision displayed a strong tendency of moving from left to right and top to bottom, but this path may not be ideal for reading a scientific paper. Readers may prefer to read the figure after it has been referenced in the main body of the paper. Ideally, a human-centered tool would be able to adapt to individual preferences, which seems within the realm of possibility for the current state of generative AI.

Next, GPT-Vision writes,

\begin{quote}
    The participants were observed at home, and encouraged to cook naturally while being filmed with a camera that didn't obstruct their movements in the kitchen.
\end{quote}

\noindent This sentence is very similar to the back half of the caption, which will become important as we read more of GPT-Vision's summary. It suddenly jumps to the beginning Section 3 (Methods), which is at the bottom of the left column:


\begin{quote}
    Below the image, the text talks about the study's setting and aim, which is to examine how voice assistants can help guide users through complex tasks, like following a recipe.
\end{quote}

However, Section 3 is not immediately below figure in the page, contrary to what it might imply by describing it immediately following the figure. It then jumps backward to the paragraph between the figure and Section 3, which is the end of a section continued from the previous page.

\begin{quote}
    The authors mention that prior work highlights the importance of voice interfaces aiding navigation in instructions by supporting multiple types of jumps, handling input conversion cycles, and supporting content-based anchors.
\end{quote}

\noindent This excerpt once again closely resembles the original text, but ``handling input conversion cycles'' has no meaning in this context. It seems like a misreading or misinterpretation of ``handle implicit conversation cues.''

The closing sentence about the left column features a close paraphrase of the last sentence of section 2.4.

\begin{quote}
    This study focuses on the challenges faced and offers audio-first task guidance and suggestions for overcoming these challenges.
\end{quote}

The ``summary'' of the left column covers very little of it: it paraphrases a the figure caption, a sentence from Section 3, and a couple of sentences from Section 2.4 while omitting key information like the research questions and study design choices. Furthermore, GPT-Vision may not be performing a language task after all---it may be picking visual details to relay, much like it picks elements of diagrams to describe. It also collapsed the entire left column into one section even though scientific papers are not meant to be read that way. Vision models that present syntheses of text material should make sure to represent the full scope well, at the risk of readers unknowingly missing crucial points.

\subsection{Respecting Boundaries}
\label{sec:boundaries}

\begin{wrapfigure}{R}{0.35\textwidth}
    \includegraphics[width=0.35\textwidth]{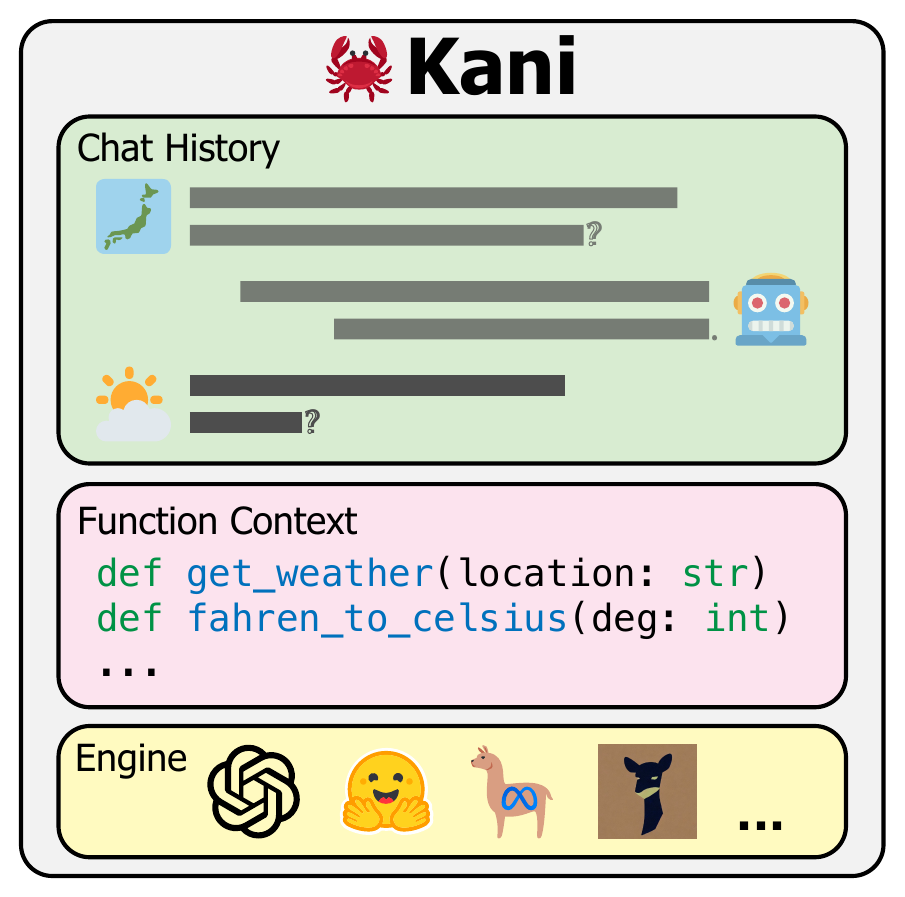}
    \caption{Excerpt from D1, an overview of ``Kani'' a framework for building chat-based LLM applications \mbox{\citep{zhu_kani_2023}}.}
    \label{fig:d1_kani_kani}
\end{wrapfigure}

GPT-Vision adeptly described individual elements in many generated passages, but it diminished when speaking of overlapping elements in D1. D1 is an overview of Kani, a framework for building chat-based applications \citep{zhu_kani_2023}. The diagram shows a cartoon avatar of a user talking with a chatbot that is powered by Kani. Kani contains three components, which are represented as three rectangles. GPT-Vision described the Kani square as follows:

\begin{quote}
    The top section of this [the Kani square] shows a chat history window. The window has the name `Kani' at the top and next to the name there is a red crab icon... Just below the chat history window, there is a section called `Function Context.' ... It's in a box with rounded edges and a light yellow background. (D1 desc)
\end{quote}

GPT-Vision blurs a few details here. The label ``Kani,'' and the crab, \textit{does} exist at top of the Kani square, but it is \textit{outside} the rectangle labeled ``Chat History.'' GPT-Vision correctly states that chat history and function context are the top two rectangles in that order, but it stated the wrong color. Function context is actually pink. This was not the only time GPT-Vision seemed to mix up nearby elements (see P1). Identifying absolute positions is a great start, but vision models need to interpret structural relationships well to represent the full range of images properly.

\subsection{Spatial Relationships}
\label{sec:spatial}

\begin{wrapfigure}{r}{0.35\textwidth}
    \includegraphics[width=0.35\textwidth]{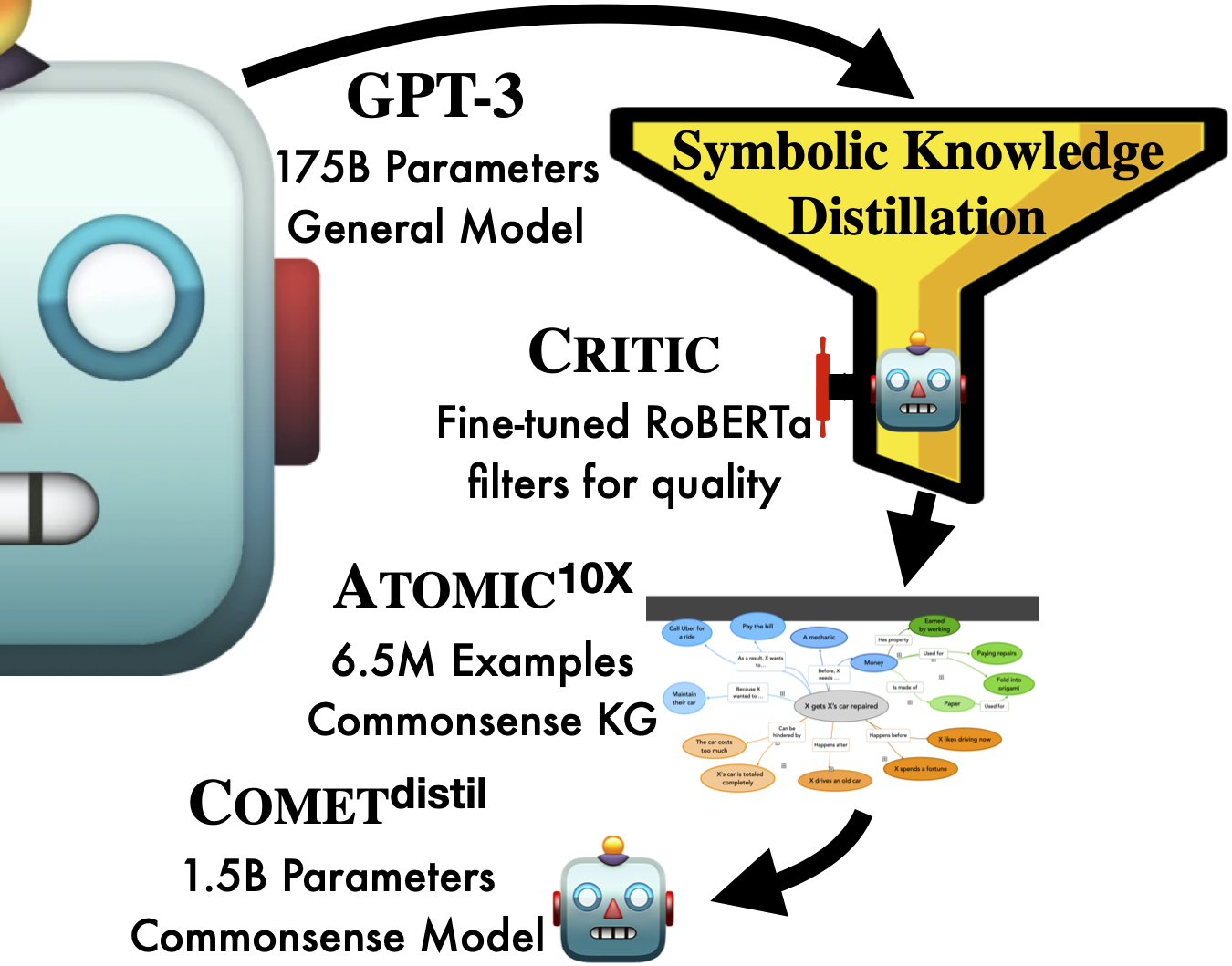}
    \caption{Symbolic knowledge \mbox{distillation} (D3) \citep{west_symbolic-knowledge-distillation_2022}.}
    \label{fig:d3_spatial}
\end{wrapfigure}


One of GPT-Vision's most consistent successes was in describing the positions of elements in an image. When describing a complex diagram about symbolic knowledge distillation by \citet{west_symbolic-knowledge-distillation_2022}, GPT-Vision accurately stated where each piece of the diagram was located:

\begin{quote}
    [In] the top left corner, there's a cartoonish depiction of a robot [...] Next to this robot character, in the top center of the image, there's some text that reads ``GPT-3'' with three bullet points below it saying ````175B Parameters'''', ````General Model'''' (D3 desc)
\end{quote}

\noindent Even though the ``three bullet points'' do not exist in the image, GPT-Vision described the positions of the elements, and the elements themselves, very well.

One of GPT-Vision's frequent weaknesses, however, was in describing the relationships between these elements. ``GPT-3,'' ``175B Parameters,'' and ``General Model'' are not arbitrary floating pieces of text; they are labels that describe what the robot represents. GPT-Vision did manage to present this detail in the ``alt'' prompt:

\begin{quote}
    On the upper left corner, there is an illustration of a robot, representing GPT-3 which has 175 billion parameters and is labeled as a General Model. (G4 alt)
\end{quote}

GPT-Vision often described the same images differently when responding to different prompts, amplifying the challenge of finding coherent patterns in its behavior. We should investigate GPT-Vision with a larger number of samples, experiment with controlled changes in prompts, and unveil the cognitive and neural structures beneath the behavior to learn more.

P1 alt and desc, P2 desc, D2 alt and desc, D3 alt and desc, G2 desc, and G3 desc, F2 alt and desc, F1 desc, F3 desc, F4 alt and desc, F5 desc, C2 desc, and M1 desc contain mentions of spatial relationships.

\subsection{Graphic Misinterpretations}

\begin{figure}[h]
    \centering
    \begin{subfigure}[t]{0.45\textwidth}
        \includegraphics[width=\textwidth, left]{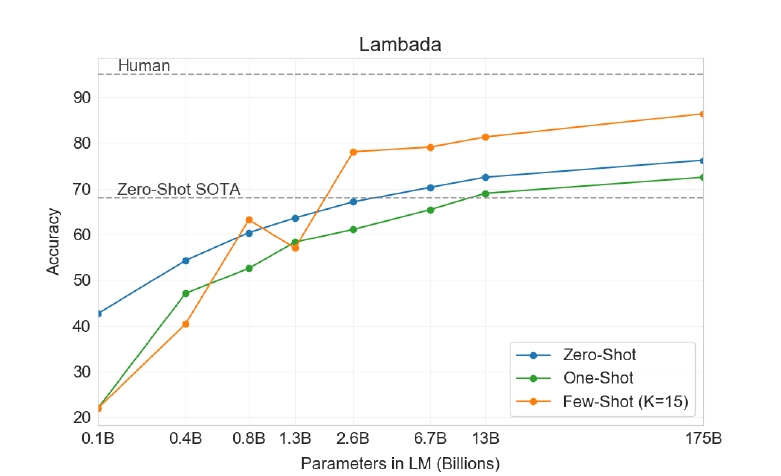}
        \caption{G3, a line graph portraying the accuracy of zero-shot, one-shot, and few-shot prompting on the LAMBADA dataset as language model size increases.}
        \label{fig:g3}
    \end{subfigure}%
    ~
    \begin{subfigure}[t]{0.45\textwidth}
        \includegraphics[width=\textwidth, left]{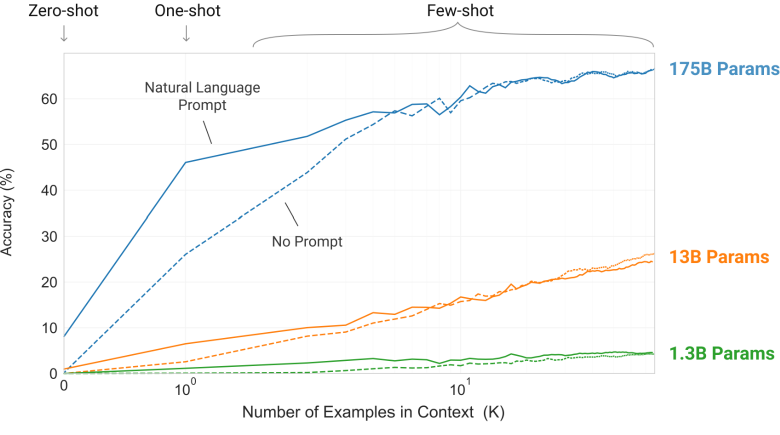}
        \caption{G4, a line graph suggesting that increased model size leads to improved in-context learning abilities.}
        \label{fig:g4}
    \end{subfigure}
    \caption{Line graphs from \citet{brown_gpt3-few-shot_2020}.}
    \label{fig:line-graphs}
\end{figure}

GPT-Vision struggled with graphs, like G3 and G4. Both are line graphs sourced from the publication introduced GPT-3, a text-only predecessor to GPT-Vision \citep{brown_gpt3-few-shot_2020}. G3 (Figure \ref{fig:g3} shows the accuracy of zero-shot, one-shot, and few-shot prompting as GPT-3 increases from 0.1 to 175 billion parameters.

G4 (Figure \ref{fig:g4}) is more complex and shows the accuracy of 1.3 billion--, 13 billion--, and 175 billion--parameter versions of GPT-3 as the number of in-context examples grows from 0 to 32. Each model size is represented by two lines: a solid line for a ``Natural Language Prompt'' and a dashed line for ``No Prompt.'' The graph contains free-floating labels for prompt style and model size as opposed to a legend, like in G3.

\paragraph{Axes} GPT-Vision described the x- and y-axes of each line graph moderately well, except that it consistently underestimated the bounds of the axes depending on the labels. For example, the y-axis in G4 is labeled from 0 to 60 in increments of 10, but the line itself extends to 70 without a tick label for $y=70$:

\begin{quote}
    The y-axis, or vertical axis, is labeled ``Accuracy (\%)'' and has a linear scale ranging from 0 to 60. (G4 desc)
\end{quote}

This seems to be a stylistic trend because the bar graph (G1) and both line graphs (G3, G4) omit the tick label for the greatest value on the y-axis. GPT-Vision mistook the bounds of an axis when describing all three of these graphs, which was particularly precarious when the data went beyond the printed bounds (G3, G4). GPT-Vision appeared to have a bias toward text in an image when it incorporated adversarial labels into its output (see Section \ref{sec:typographical-influence}). Future work in ``artificial cognition'' to expose what GPT-Vision pays attention to can help mitigate this weakness.

GPT-Vision made a subtler text-based error when describing the x-axis in G4:

\begin{quote}
    The x-axis... has a logarithmic scale, starting at 10\^{}0 and increasing to 10\^{}1. (G4 desc)\end{quote}

\noindent The x-axis is labeled with ``$0$,'' ``$10^0$,'' and ``$10^1$,'' reminiscent of a logarithmic scale, but ``$10^1$'' is further from ``$10^0$'' than ``$10^0$'' is from ``$0$.'' These values would be equally spaced on a true logarithmic scale. Successfully reading axes partially requires the ability to judge visual distance because we estimate values on a graph by examining how close a point is to a value on a number line. GPT-Vision has already shown a good start in describing positions of elements in an image (see Section \ref{sec:spatial}), inspecting how well GPT-Vision describes the amount of space between two points is a natural next step.

\paragraph{Data trends} GPT-Vision imprecisely represented data trends in both line graphs. G3, for example, displays three solid lines with circle markers for ``Zero-Shot'' (blue), ``One-Shot'' (green), and ``Few-Shot (K=15)'' (orange) prompting. It described the lines qualitatively with some clarity:

\begin{quote}
    The third line, depicted in blue and labeled ``Zero-Shot'', appears to be an upward leaning curve... The fourth line, represented in green and labeled ``One-Shot'', is similar to the third but starts at a slightly higher accuracy... Lastly, an orange line labeled ``Few-Shot (K=15)''... increases quite sharply... (G3 desc)\end{quote}

The Zero-Shot and One-Shot curves do look similar to each other, starting lower and rising gently. The One-Shot line, however, starts at a \textit{substantially lower} accuracy (about 20\%) than Zero-Shot (about 40\%).

\paragraph{Numerical estimates} GPT-Vision also imprecisely estimated the starting and ending values of each curve: it noted Zero-Shot as ranging from 30\% to 60\% (closer to 40\%--75\%), One-Shot from 40\% to 70\% (closer to 20\%--70\%), and Few-Shot (K=15) from 35\% to 90\% (closer to 20\%--85\%). It stated that Few-Shot (K=15) ``surpass[ed] the One-Shot accuracy at around 2.6 billion parameters,'' which is inaccurate as well---Few-Shot surpassed One-Shot much earlier, between 0.4 and 0.8 billion parameters.

Similar behavior occurred in both generated passages for G4 as well. The line representing GPT-3 1.3B starts at 0\% accuracy and remains nearly flat, but GPT-Vision described it as

\begin{quote}
    barely rising above 10\% accuracy as the number of examples increases (G4 desc)
\end{quote}

\noindent which implies that the model achieved at least 10\% accuracy. However, G4 shows GPT-3 1.3B remaining well under the 10\% grid line, which the response to the ``alt'' prompt actually describes appropriately.

\begin{quote}
    1.3 billion parameter model has the least accuracy, remaining below 10\%... (G4 alt)
\end{quote}

These mixed insights hint at GPT-Vision's emerging graph-reading abilities, especially when it described the shapes of the lines in G3 and G4. Teaching GPT-Vision to read axes properly would allow it to make deeper insights about complex data, and maybe even uncover some unnoticed trends.

\subsection{Writing the Math Out}
\label{sec:math}

GPT-Vision generated numerous errors when reproducing mathematical text, from misprinting ``$1^5$'' as ``1\^{}5\^{}2'' (T3 desc), to misrepresenting ``$\left(\alpha - \frac{(1 - \alpha) 2mw(\left\lceil m \right\rceil) }{k-1} \right)$'' as ``$(\alpha - 1 / (2k - 2))$'' (C1 alt). These errors can have drastic consequences if they are not corrected or verified. In addition, GPT-Vision often produced \LaTeX-style, some of which would not have compiled (T3 desc, M1 and desc, and C1 alt and desc). 

Besides the downstream challenge that the end-user's system may not render \LaTeX, many of the \LaTeX-style reproductions were wrong. GPT-Vision sometimes omitted subscripts or misprinted them (which could be attributed to low resolution). It was particularly prone to error when a subscript was longer than one character. In \LaTeX, a subscript starts with an underscore followed by the characters to be subscripted. A single character can be written alone, like ``x\_i = $x_i$,'' but multiple characters need to wrapped in curly braces. In all cases except one, GPT-Vision missed this distinction. For example, it reproduced ``$\alpha_{xy}$'' as ``a\_xy,'' which would have compiled to ``$a_xy$'' (C1 alt). These errors were internally consistent, so GPT-Vision referred to the same values in the same way within each passage.

The one multi-character subscript GPT-Vision reproduced correctly was from this equation: 

\begin{equation*}
    p(y_i | y1, ..., y_{i-1}, \mathbf{x}) = g(y_{i-1}, s_i, c_i),
\end{equation*}

\noindent which it wrote as

\begin{center}
    p(y\_i$|$y\_1, ..., y\_\{i-1\}, x) = g(y\_\{i-1\}, s\_i, c\_i) (M1 desc)
\end{center}

\noindent Given the frequency of its \LaTeX \ errors and the ubiquity of this expression as a conditional probability for a recurrent neural network, we should avoid assuming that GPT-Vision ``understood'' how to print it correctly. Assessing the underlying abilities of closed-source models is a common challenge since we cannot verify the training data, but our behavioral analysis seems to suggest that this success is coincidental.

Powerful generative AI models like GPT-Vision could go even further by describing mathematical text in natural language instead of solely reproducing it as is. GPT-Vision already showed good performance when describing a pseudocode algorithm (see Section \ref{sec:code}). With adjustments, GPT-Vision has great potential to advance learning, accessibility, and inclusion.

\subsection{Counting Errors}
\label{sec:number-sense-counting}
\begin{wrapfigure}{r}{0.45\textwidth}
\renewcommand{\arraystretch}{1.2}
\centering
\begin{tabular}{ccc}
\toprule
Original & \# & GPT-V \\ 
\midrule
\boxes{4}{5}     & 4       & 3    \\
\boxes{4}{5}     & 4       & 3    \\
\boxes{2}{5}     & 2       & 4    \\
\boxes{3}{5}     & 3       & 4    \\
\boxes{1}{5}     & 1       & 4    \\
\boxes{2}{5}     & 2       & 3    \\
\boxes{4}{5}     & 4       & 3    \\
\boxes{2}{5}     & 2       & 5    \\
\boxes{5}{5}     & 5       & 3    \\
\boxes{4}{5}     & 4       & 4    \\
\boxes{2}{5}     & 2       & 5    \\
\boxes{3}{5}     & 3       & 3    \\
\bottomrule
\end{tabular}
\caption{The skill levels from T1 (Original) with their true counts (\#) and GPT-Vision's interpretation \citep{hwang_rewriting_2023}.}
\label{tab:counting-boxes}
\end{wrapfigure}

Counting objects was another frequent source of error, with GPT-Vision miscounting on 10 of 21 images. For example, T1 shows a table of participants from a cooking study. One column, labeled ``Self-Rated Skill,'' lists how the participants rated their cooking abilities on a five-point Likert scale. Instead of listing the number, the table presents the skill levels with a sequence of five boxes. The number of large ``filled'' boxes represents the participant's skill level out of five.

For example, this reproduction {\boxes{3}{5}} shows three large boxes followed by two small boxes, so it represents a skill level of three out of five. GPT-Vision counted ten out of twelve of these boxes incorrectly when responding to the ``desc'' prompt (it did not count at all for the ``alt'' prompt) (see Table \ref{tab:counting-boxes}).

The responsibility for LLMs to handle numbers well is unclear. Some have argued that LLMs should be given a calculator rather than be trained to calculate \citep{andor_giving-bert-calculator_2019}. The mechanism for numerical reasoning may vary, but number sense will remain an important capability for describing images.

P1 alt and desc, G1 alt, G2 alt and desc, G3 desc, T1 desc, T2 desc, T3 alt and desc, F2 desc, F4 desc, and C2 desc contain counting errors.

\subsection{(Lack of) Logo Recognition}
\label{sec:logos}

\begin{figure}[h]
    \centering
    \includegraphics[width=0.6\textwidth]{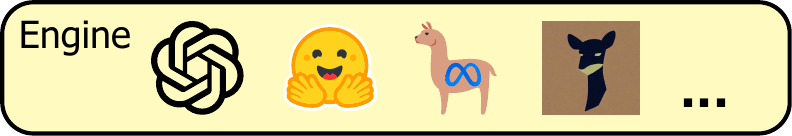}
    \caption{The Engine section of D1 \citep{zhu_kani_2023}.}
    \label{fig:d1_kani_engine}
\end{figure}

GPT-Vision did not recognize the three logos displayed in the ``Engine'' section at the bottom of D1. These logos were supposed to represent the language models that Kani supports:

\begin{enumerate}
    \item OpenAI (a circular logo resembling three intertwined chain links) \citep{openai_openai_2023},
    \item Hugging Face (a yellow emoji-like happy face with open hands) \citep{huggingface_hugging-face_2023},
    \item LLaMA (not an official logo, a brown cartoon llama with the Meta logo) \citep{touvron_llama-open-foundation_2023},
    \item Vicuna (head of a cartoon vicuna, which has a tall neck and pointy ears) \citep{thevicunateam_vicuna-open-source-chatbot_2023}.
\end{enumerate}

\noindent which GPT-Vision described in the desc passage as

\begin{enumerate}
    \item ``a caduceus [two serpents twisted around a staff] with only one snake'' \citep{wikipedia_caduceus_2023},
    \item ``a yellow smiley face,''
    \item ``a flamingo,'' and
    \item ``a letter `Y' with what looks like animal ears on top.''
\end{enumerate}

\noindent Besides mistaking the llama for a flamingo, GPT-Vision's descriptions of the engine icons are not far off, but not recognizing the logos themselves obscured the point of this part of the diagram. 

\subsection{Color Blindness}
\label{sec:colors}

\begin{table}[h]
    \small
    \centering
    \begin{tabular}{ll|l>{\raggedright}p{4.5cm}}
        \toprule
        \multicolumn{2}{c|}{\textbf{Original Colors}} & \multicolumn{2}{c}{\textbf{GPT-Vision's Interpreted Colors}} \tabularnewline
        \multicolumn{1}{c}{Color} & \multicolumn{1}{c|}{Category} & \multicolumn{1}{c}{Color}  & \multicolumn{1}{c}{Category} \tabularnewline
        \midrule
        \singlebox{grammar} green               & grammar        & \singlebox{irrelevant} blue           & grammar errors                             \tabularnewline
        \singlebox{repetition} orange           & repetition     & \singlebox{grammar} green           & repetitions                                \tabularnewline
        \singlebox{irrelevant} blue           & irrelevant     &  \singlebox{purple} purple            & irrelevance                               \tabularnewline
        \singlebox{contra-sent} pink            & contradicts\_sentence  & \multirow{2}{*}{\tolbox{common-sense}} \multirow{2}{*}{yellow}   & \multirow{2}{4.5cm}{contradictions with sentence context or knowledge}                     \tabularnewline
        \singlebox{contra-know} light green     & contradicts\_knowledge    & & \tabularnewline
         \multirow{2}{*}{\singlebox{common-sense} yellow}         & \multirow{2}{*}{common\_sense}        & \multirow{2}{*}{\singlebox{light-purple} light purple}     & common sense and \tabularnewline
        &&& coherence errors \tabularnewline
        \singlebox{coreference} tan             & coreference       & \singlebox{repetition} orange         & coreference errors                        \tabularnewline
        \singlebox{generic} gray                & generic           & \singlebox{red} red        & generic                                   \tabularnewline
        \singlebox{other} green                 & other             & \singlebox{generic} gray        & other errors                               \tabularnewline
        \bottomrule
    \end{tabular}
    \caption{The legend from G2 with GPT-Vision's interpretation (desc) \citep{dugan_roft-analysis_2023}.}
    \label{tab:colors}
\end{table}

As mentioned in \citet{openai_gptvision_2023}, GPT-Vision consistently failed to recognize colors. This was especially apparent when it described G2, a plot of pie charts with nine color-coded categories (see Table \ref{tab:colors}). The legend in G2 shows a vertical list of categories with their associated colors:

\begin{quote}
    (1) ``grammar'' (green), (2) ``repetition'' (orange), (3) ``irrelevant'' (blue), (4) ``contradicts\_sentence'' (pink), (5) ``contradicts\_knowledge'' (light green), (6) ``common\_sense'' (yellow), (7) ``coreference'' (tan), (8) ``generic'' (gray), and (9) ``other'' (green, repeated),
\end{quote}

\noindent but GPT-Vision reported them slightly differently (\textbf{emphasis} ours).

\begin{quote}
    (1) ``grammar errors,'' (2) ``repetition\textbf{s},'' (3) ``irrelevan\textbf{ce},'' (4+5) ``contradictions with sentence context or knowledge,'' (6+) ``commonsense \textbf{and coherence} errors,'' (7) ``coreference errors,'' and (8) ``other errors.'' (G2 desc)
\end{quote}

GPT-Vision also mislabeled most of the colors. It mistook green for blue, orange for green, blue for purple, yellow for light purple, tan for orange, gray for red, and green for gray (G2 desc). It seems to have merged ``contradicts\_sentence'' (pink) and ``contradicts\_knowledge'' (light green) into one category of the color yellow. GPT-Vision displayed similar behavior with the legend in G1 as well, suggesting that color recognition is a serious weakness. It may define colors differently than we expect or suffer from one-off errors from misaligning the color swatches with their labels. Further investigation into the source of this behavior may help us fix it.

\subsection{Quality of Alt Text}
\label{sec:quality-alt-text}

\paragraph{Length} Most of the alt text generated by GPT-Vision was about a paragraph in length, the exception being alt text for full pages that was typically much longer. This clashes with standard guidelines for alt text, which recommend a brief sentence because screen readers may impose character limits \citep{w3_images-tutorial_2022, vleguru_alt-text-video_2022}. One work in generating image descriptions for accessibility, however, found that blind/low-vision participants actually preferred longer descriptions (while sighted participants showed no clear pattern), in contrast with typical guidelines \citep{kreiss_context-matters-image_2022}. With the right adjustments, GPT-Vision's ability to generate long text can lead to detailed, valuable image descriptions.

\paragraph{Audience, content, purpose} Good alt text depends on the audience, content, and purpose of the image, so one alt text cannot necessarily fit all situations for the same image \citep{vleguru_alt-text-video_2022}. Our analysis found that GPT-Vision tended to focus too much on visual details and too little on the main ideas.  For example, when describing a full-page screenshot of \citet{hwang_rewriting_2023}, GPT-Vision wrote,

\begin{quote}
    This is an image of a research paper page titled ``Rewriting the Script: Adapting Text Instructions for Voice Interaction.'' The page contains a figure and two sections of text with bullet points. (F3 alt)\end{quote}

The details GPT-Vision chose to highlight misrepresent the likely audience and purpose of this image. The audience of such a paper is likely to be researchers in computer science or user experience design. The purpose of the page is to convey information about the study, namely the methods, technology probe, and participants. GPT-Vision instead surfaced the figure and ``two sections of text'' to the reader, giving them very little idea of the content itself. A more useful alt text may have been

\begin{quote}
    Page from a research paper titled ``Rewriting the Script: Adapting Text Instructions for Voice Interaction'' discussing part of section 3, ``METHODS,'' with a figure of the ``study setting'' and a table of ``participants.''
\end{quote}

Not all readers are the same, of course. \citet{lundgard_accessible-visualization-natural_2021} found a stark divide between blind/low-vision (BLV) and sighted readers for graphs: sighted readers appreciated a ``story'' about the data but BLV readers strongly disliked ``subjective interpretations, contextual information, or editorializing.'' BLV readers wanted a more literal description of the graph so they could interpret the data for themselves.

For BLV readers, the emphasis on visual details that GPT-Vision tended to provide may be very useful if it selects the most important details to describe. Sighted readers will need a different kind of alt text while readers with non-visual disorders like issues with long- or short-term visual memory may need another set of standards altogether. GPT-Vision showed impressive performance on a diverse set of images with just two simple prompts. It shows great promise to generate high-quality alt text for more than just scientific images.
\section{Conclusion}


In this paper we have presented a framework for a more rigorous and structured application of qualitative analysis to generative AI models. Our proposed framework not only alleviates the concerns of previous large-scale qualitative analysis work being ``unscientific'', but also allows us the opportunity to develop an alternative approach to evaluation separate from traditional benchmarks. Through our analysis we are able to identify a number of general trends in the capabilities of the newly-released GPT-Vision model such as its heavy reliance on textual information and its sensitivity to prompts. Such insights will no doubt be useful in future applications and can serve as guidelines for future areas of research.

One important caveat is that, while our analysis offers key insight on GPT-Vision's \textit{behavior} with scientific images, such insights should not be conflated with a statistical understanding of the relative frequency of these issues or a scientific explanation of why such issues occur. Even a suitable description of an image does not necessarily mean that GPT-Vision ``explained the image'' if the same information could have been hallucinated from its internal knowledge or training data. Much like in psychology, \textit{behavioral} studies cannot fully supplant \textit{cognitive} research or \textit{neuroscience}. Further investigation on the ``cognitive'' processes of LLMs, like attention and memory, and the mathematical basis of neural networks is crucial for understanding LLMs holistically.

\section*{Acknowledgments}
First and foremost, we would like to thank Liam Dugan for his tremendous support and feedback. We were also inspired by early talks with Jonathan Bragg and Doug Downey. We are also grateful for feedback from Harry Goldstein, Andrew Zhu, and Natalie Collina on GPT-Vision's descriptions of their work. Finally, we are thankful for the community at Penn NLP and Penn HCI that could make this work possible.

\selectlanguage{english}
\bibliographystyle{acl_natbib}
\bibliography{references}

\appendix

\begin{table}[]
\renewcommand{\arraystretch}{1.2}
\centering
\begin{tabular}{lc>{\raggedright}p{12cm}}
\toprule
\multicolumn{1}{c}{Type} & ID    & \multicolumn{1}{c}{Description} \tabularnewline
\midrule
\multirow{4.5}{*}{\STAB{\rotatebox[origin=c]{90}{Photo}}} & P1    & A 2x6 photo collage of various dishes labeled with their names prepared in \mbox{\citep{hwang_rewriting_2023}}. \tabularnewline
& P1.1 & The same image as P1 but with modified labels \mbox{\citep{hwang_rewriting_2023}}. \tabularnewline
& P2    & A study participant cooking in a kitchen with an Alexa Echo Dot, which is circled on the right \mbox{\citep{hwang_rewriting_2023}}. \tabularnewline
\midrule
\multirow{6}{*}{\STAB{\rotatebox[origin=c]{90}{Diagram}}} & D1     & An illustration of Kani, a framework for building applications with large language models, from the first page of \mbox{\citep{zhu_kani_2023}} \tabularnewline
& D2    & The transformation of a written recipe for audio delivery by editing the original text \mbox{\citep{hwang_rewriting_2023}}. \tabularnewline
& D3    & A complex, abstract representation of symbolic knowledge distillation with emojis, arrows, text labels, and other visual details \mbox{\citep{west_symbolic-knowledge-distillation_2022}}. \tabularnewline
\midrule
& G1    & A plot of three bar graphs, each displaying two groups of four differently colored columns with error bars \mbox{\citep{dugan_roft-analysis_2023}}. \tabularnewline
& G2    & A 3x3 plot of 9 pie charts representing 9 color-coded categories \mbox{\citep{dugan_roft-analysis_2023}}. \tabularnewline
& G3    & A line graph displaying three color-coded data lines and two dashed horizontal benchmark lines \mbox{\citep{brown_gpt3-few-shot_2020}}. \tabularnewline
\multirow{-6}{*}{\STAB{\rotatebox[origin=c]{90}{Graph}}} & G4     & A line graph similar to G3, but with three pairs lines (dashed and solid) and additional text labels on the plot \mbox{\citep{brown_gpt3-few-shot_2020}}. \tabularnewline
\midrule
& T1    & A 13x4 table (including header) containing text and sequences of boxes graphically representing Likert scales \mbox{\citep{hwang_rewriting_2023}}. \tabularnewline
& T1.1  & The same table as T1, but with the caption included beneath it. \tabularnewline
& T2    & A 6x10 (including 2 rows for the header) table of performance metrics; some columns are merged  \mbox{\citep[Table 18]{och_systematic-comparison-various_2003}}. \tabularnewline
\multirow{-4.5}{*}{\STAB{\rotatebox[origin=c]{90}{Table}}} & T3     & A 16x6 table (including 2 rows for the header) of model training schemes for varying corpus sizes; some rows are merged \mbox{\citep[Table 4]{och_systematic-comparison-various_2003}}. \tabularnewline
\bottomrule
\end{tabular}
\caption{Images of figures used in our analysis.}
\label{tab:data_figures}
\end{table}

\begin{table}
\renewcommand{\arraystretch}{1.2}
\centering
\begin{tabular}{cc>{\raggedright}p{12cm}}
\toprule
\multicolumn{1}{c}{Type} & ID   &  \multicolumn{1}{c}{Description} \tabularnewline
\midrule
& F1    & The first page of a research publication with the title, authors, two columns of text, and metadata \mbox{\citep{hwang_rewriting_2023}}. \tabularnewline
& F2    & The full page of a research publication that includes P1 and its caption spanning the top half followed by two columns of text \mbox{\citep{hwang_rewriting_2023}}. \tabularnewline
& F3    & The full page of a research publication displaying P2 in the top of the left column of text and T1 in the top of the right \mbox{\citep{hwang_rewriting_2023}}. \tabularnewline
& F4    & A full page of a research publication with a large, text-based table covering the top two-thirds and some text in two columns beneath it \mbox{\citep{hwang_rewriting_2023}}. \tabularnewline
\multirow{-8.5}{*}{\STAB{\rotatebox[origin=c]{90}{Full page}}} & F5  &   A full page of a research publication in one-column format beginning with two brief side-by-side snippets of Haskell code \mbox{\citep{goldstein_pl-reflecting-random_2023}}. \tabularnewline
\midrule
& C1    & A brief pseudocode algorithm featuring a nested for loop, an if statement, and some mathematical text \mbox{\citep{collina_algorithm-dynamic-weighted-matching_2021}}. \tabularnewline
& C2    & A page-long proof from a research publication on programming languages that includes Haskell code \mbox{\citep{goldstein_pl-reflecting-random_2023}}. \tabularnewline
\multirow{-3.7}{*}{\STAB{\rotatebox[origin=c]{90}{Code}}} & C3     & A page-long excerpt of Python code defining a class and a few instance methods for a chat-based application \mbox{\citep{zhu_kani_2023}}. \tabularnewline
\midrule
\multirow{5}{*}{\STAB{\rotatebox[origin=c]{90}{Math}}} & M1    & An excerpt from a machine learning research publication introducing a new model architecture with mathematical equations \mbox{\citep{bahdanau_seq2seq-attention_2016}}. \tabularnewline
& M2    & The definition of an algorithmic theorem followed by its proof, featuring bullet points and mathematical representations of abstract concepts \mbox{\citep{collina_algorithm-dynamic-weighted-matching_2021}}. \tabularnewline
\bottomrule
\end{tabular}
\caption{Images of full pages and special text (code and math) used in our analysis.}
\label{tab:data_full-special}
\end{table}

\end{document}